\documentclass[10pt,twocolumn,letterpaper]{article}

\usepackage{cvpr}
\usepackage{times}
\usepackage{epsfig}
\usepackage{graphicx}
\usepackage{amsmath}
\usepackage{amssymb}

\usepackage{multirow}
\usepackage{makecell}
\usepackage{caption}
\usepackage{subcaption}
\usepackage{algorithm2e}
\usepackage{algpseudocode}
\usepackage{float}
\usepackage{url}

\usepackage{color}
\usepackage{xcolor,colortbl}
\usepackage{framed}
\usepackage{enumitem}
\usepackage{booktabs}

\definecolor{shadecolor}{rgb}{0.9,0.9,0.9}
\definecolor{Gray}{gray}{0.9}

\usepackage[pagebackref=true,breaklinks=true,letterpaper=true,colorlinks,bookmarks=false]{hyperref}

\cvprfinalcopy 

\def\cvprPaperID{556}

\ifcvprfinal\fi

\begin{document}

\title{Large-Scale Long-Tailed Recognition in an Open World}

\author{Ziwei Liu$^{1,2}$\thanks{Equal contribution.} ~~~ Zhongqi Miao$^{2*}$ ~~~ Xiaohang Zhan$^{1}$ ~~~ Jiayun Wang$^{2}$ ~~~ Boqing Gong$^{2}\thanks{Work done in part at Tencent AI Lab.}$ ~~~ Stella X. Yu$^{2}$ \\
$^{1}$ The Chinese University of Hong Kong ~~~~~~~~~ $^{2}$ UC Berkeley / ICSI\\
{\tt\small \{zwliu,zx017\}@ie.cuhk.edu.hk, \{zhongqi.miao,peterwg,stellayu\}@berkeley.edu, bgong@outlook.com}
}

\maketitle
\thispagestyle{empty}

\begin{abstract}

Real world data often have a long-tailed and open-ended distribution. A practical recognition system must classify among majority and minority classes, generalize from a few known instances, and acknowledge novelty upon a never seen instance.  We define Open Long-Tailed Recognition (OLTR) as learning from such naturally distributed data and optimizing the classification accuracy over a balanced test set which include head, tail, and open classes.

OLTR must handle imbalanced classification, few-shot learning, and open-set recognition in one integrated algorithm, whereas existing classification approaches focus only on one aspect and  deliver poorly over the entire class spectrum.  The key challenges are how to share visual knowledge between head and tail classes and how to reduce confusion between tail and open classes.

We develop an integrated OLTR algorithm that maps an image to a feature space such that visual concepts can easily relate to each other based on a learned metric that respects the closed-world classification while acknowledging the novelty of the open world.  Our so-called dynamic meta-embedding combines a direct image feature and an associated memory feature, with the feature norm indicating the familiarity to known classes.
On three large-scale OLTR datasets we curate from object-centric ImageNet, scene-centric Places, and face-centric MS1M data, our method consistently outperforms the state-of-the-art.  Our code, datasets, and models enable future OLTR research and are publicly available at \url{https://liuziwei7.github.io/projects/LongTail.html}.
    
\end{abstract}



\section{Introduction}

Our visual world is inherently long-tailed and open-ended: 
The frequency distribution of visual categories in our daily life is long-tailed~\cite{reed2001pareto}, with a few common classes and many more rare classes, and we constantly encounter new visual concepts as we navigate in an open world.

\begin{figure}[t]
  \centering
  \includegraphics[width=0.49\textwidth]{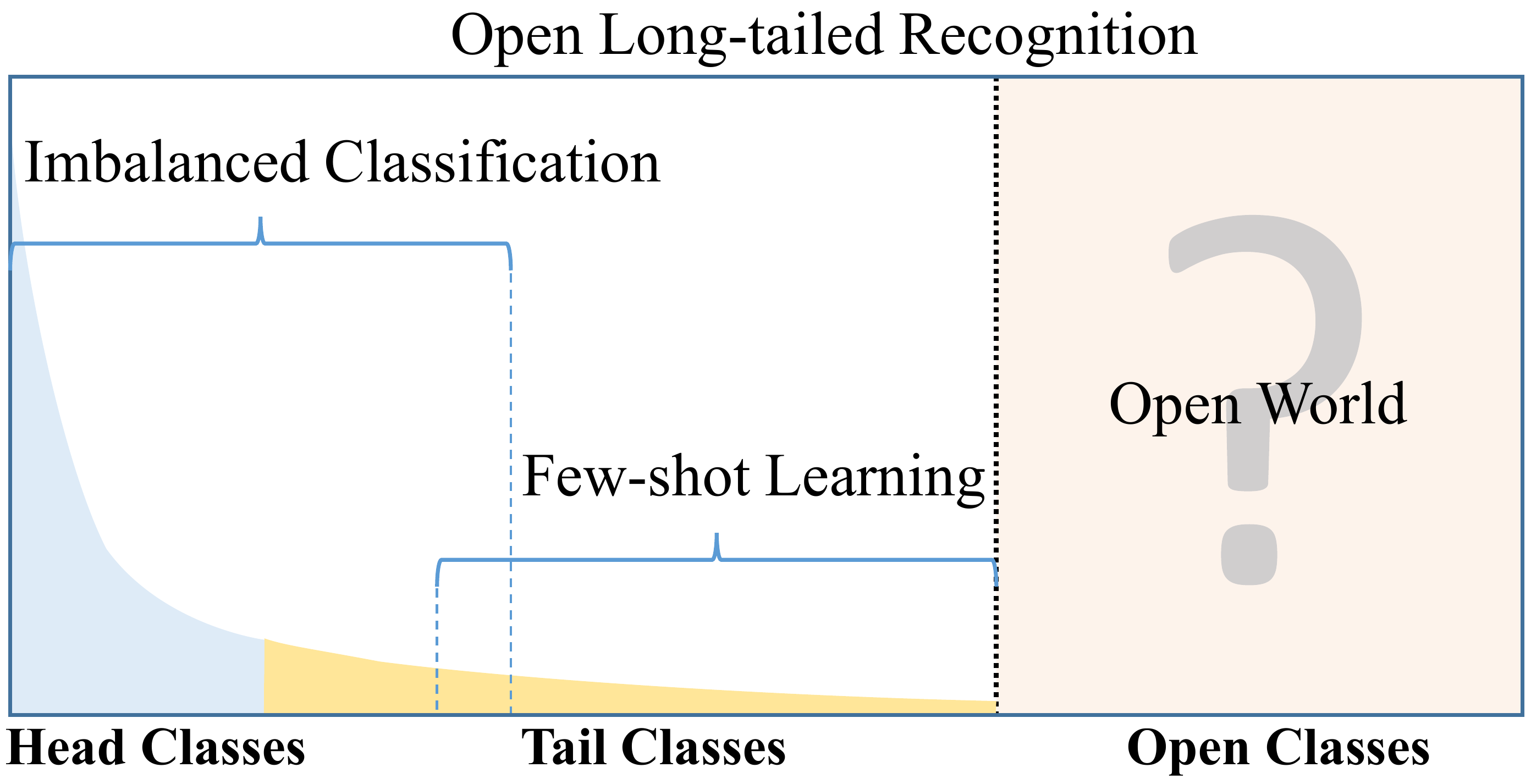}
\caption{Our task of open long-tailed recognition must learn from long-tail distributed training data in an open world and  deal with imbalanced classification, few-shot learning, and open-set recognition over the entire spectrum.}
  \label{fig:intro}
\end{figure}

\begin{table*}[t]
    \centering
    \resizebox{1.0\textwidth}{!}{%
    \begin{tabular}{l|c|c|c|c|c}
    \Xhline{1pt}
    {\bf Task Setting} & {\bf Imbalanced Train/Base Set} &  {\bf \#Instances in Tail Class } & {\bf Balanced Test Set} & {\bf Open Class} & {\bf Evaluation: Accuracy Over ?} \\ \hline \hline
    Imbalanced Classification & $\checkmark$ & 20$\sim$50  & $\times$ & $\times$ &  all classes \\ \hline
    Few-Shot Learning & $\times$ & 1$\sim$20 & $\checkmark$ & $\times$ &   novel classes \\ \hline
    Open-Set Recognition & $\times$ & N/A & $\checkmark$ & $\checkmark$ & all classes \\ \hline
    \bf Open Long-Tailed Recognition & $\checkmark$ & 1$\sim$20  & $\checkmark$ & $\checkmark$ & all classes \\
    \Xhline{1pt}
    \end{tabular}}
    \caption{Comparison between our proposed OLTR task and related existing tasks.} \label{tab:comparison}
\end{table*}



 While the natural data distribution contains head, tail, and open classes (Fig. \ref{fig:intro}), existing classification approaches focus mostly on the
head~\cite{deng2009imagenet, lin2014microsoft}, the tail~\cite{vinyals2016matching, lake2015human}, often in a closed setting~\cite{wang2017learning, miao2018comparison}.  Traditional deep learning  models are good at capturing the big data of head classes ~\cite{krizhevsky2012imagenet, he2016deep}; more recently, few-shot learning methods have been developed for the small data of tail classes ~\cite{snell2017prototypical, hariharan2017low}.


We formally study {\it Open Long-Tailed Recognition} (OLTR) arising in natural data settings.  A practical system shall be able to classify among a few common and many rare categories, to generalize the concept of a single category from only a few known instances, and to acknowledge novelty upon an instance of a never seen category.  We define OLTR as learning from long-tail and open-end distributed data and evaluating the classification accuracy over a balanced test set which include head, tail, and open classes in a continuous spectrum (Fig. \ref{fig:intro}). 

OLTR must handle not only imbalanced classification and few-shot learning in the closed world, but also open-set recognition with one integrated algorithm (Tab. \ref{tab:comparison}).
Existing classification approaches tend to focus on one aspect and deliver poorly over the entire class spectrum.

The key challenges for OLTR are tail recognition robustness and open-set sensitivity:
 As the number of training instances drops from thousands in the head class to the few in the tail class, the recognition accuracy should maintain as high as possible; on the other hand, as the number of instances drops to zero in the open set, the recognition accuracy relies on the sensitivity to distinguish unknown open classes from known tail classes.  
 
 An integrated OLTR algorithm should tackle the two seemingly contradictory aspects of recognition robustness and recognition sensitivity on a continuous category spectrum.  To increase the recognition robustness, it must share visual knowledge between head and tail classes;  to increase recognition sensitivity, it must reduce the confusion between tail and open classes.


We develop an OLTR algorithm that maps an image to a feature space such that visual concepts can easily relate to each other based on a learned metric that respects the closed-world classification while acknowledging the novelty of the open world.  

Our so-called {\it dynamic meta-embedding}
handles tail recognition robustness by combining two components:  a direct feature computed from the input image, and an induced feature associated with the visual memory. 
{\bf 1)}
Our direct feature is a standard  embedding that gets updated from the training data by stochastic gradient descent over the classification loss.  The direct feature lacks sufficient supervision for the rare tail class.  
{\bf 2)}
Our memory feature is inspired by meta learning methods with memories~\cite{vinyals2016matching, duan2016rl, ba2016using} to augment the direct feature from the image.
A visual memory holds discriminative centroids of the direct feature. We learn to retrieve a summary of memory activations from the direct feature, combined into a meta-embedding that is enriched particularly for the tail class.

Our dynamic meta-embedding handles open recognition sensitivity by dynamically calibrating the meta-embedding with respect to the visual memory.
The embedding is scaled inversely by its distance to the nearest centroid: The farther away from the memory, the closer to the origin, and the more likely an  open set instance.
We also adopt \emph{modulated attention}  ~\cite{wang2017non} to encourage the head and tail classes to use different sets of spatial features.
As our meta-embedding relates head and tail classes,  our modulated attention maintains discrimination between them. 

We make the following major contributions.
{\bf 1)}
We formally define the OLTR task, which learns from natural long-tail and open-end distributed data and optimizes the overall accuracy over a balanced test set.  It provides a comprehensive and unbiased evaluation of visual recognition algorithms in practical settings.
{\bf 2)}
We develop an integrated OLTR algorithm with dynamic meta-embedding.  It handles tail recognition robustness by relating visual concepts among head and tail embeddings, and it handles open recognition sensitivity by dynamically calibrating the embedding norm with respect to the visual memory. 
{\bf 3)}
We curate three large OLTR datasets according to a long-tail distribution from existing representative datasets: object-centric ImageNet, scene-centric MIT Places, and face-centric MS1M datasets.  We set up benchmarks for proper OLTR performance evaluation. 
{\bf 4)}
Our extensive experimentation on these OLTR datasets demonstrates that our method consistently outperforms the state-of-the-art.  

Our code, datasets, and models are publicly available at 
\url{https://liuziwei7.github.io/projects/LongTail.html}.
Our work fills the void in practical benchmarks for imbalanced classification, few-shot learning, and open-set recognition, enabling future research that is directly transferable to real-world applications.

\begin{figure*}[t]
  \centering
  \includegraphics[width=0.9\textwidth]{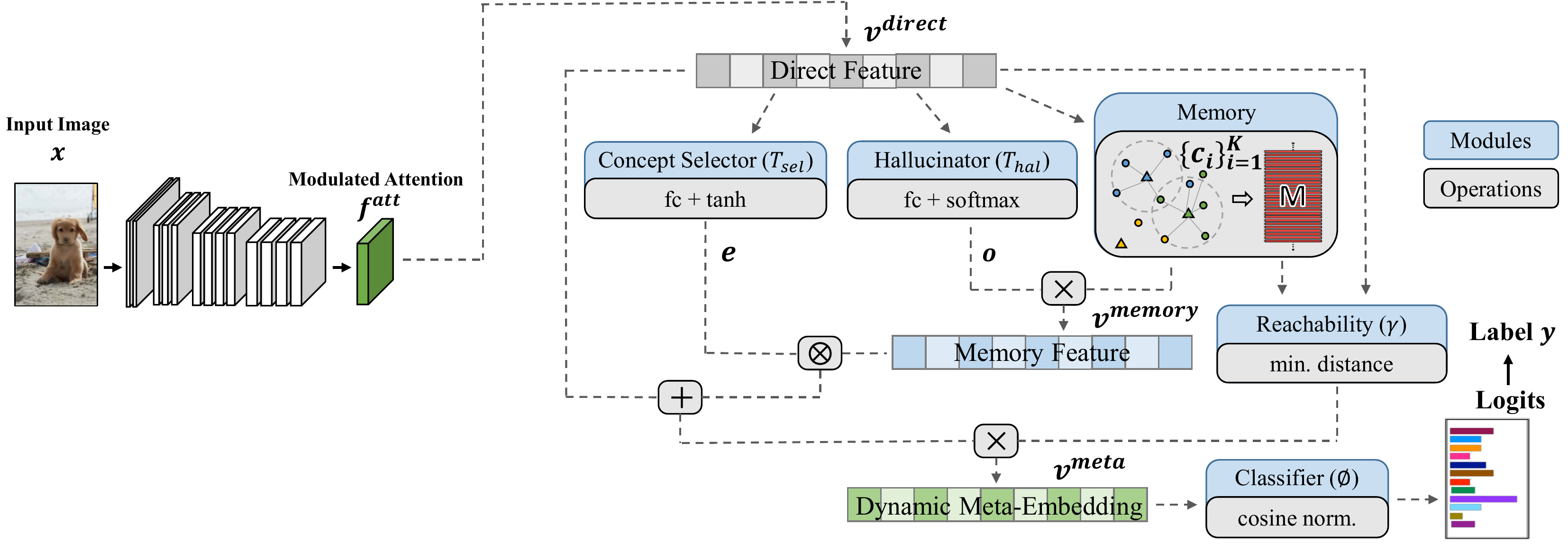}
   \caption{\textbf{Method overview.} There are two main modules: dynamic meta-embedding and modulated attention. The embedding relates visual concepts between head and tail classes, while the attention discriminates between them. The {\emph reachability} separates tail and open classes.}
  \label{fig:pipeline}
  \vspace{-6pt}
\end{figure*}


\section{Related Works}

 
While OLTR has not been defined in the literature, there are three closely related tasks which are often studied in isolation: imbalanced classification, few-shot learning, and open-set recognition. Tab.~\ref{tab:comparison} summarizes their differences.

\vspace{2pt}
\noindent
\textbf{Imbalanced Classification.}
%
Arising from long-tail distributions of natural data, it has been extensively studied ~\cite{salakhutdinov2011learning, zhu2014capturing, bengio2015the, liu2015deep, zhu2016we, ouyang2016factors, liu2016deepfashion, van2017devil, cui2018large}.
Classical methods include under-sampling head classes, over-sampling tail classes, and data instance re-weighting. We refer the readers to~\cite{he2008learning} for a detailed review.
Some recent methods include \emph{metric learning}~\cite{huang2016learning, oh2016deep}, \emph{hard negative mining}~\cite{dong2017class, lin2017focal}, and \emph{meta learning}~\cite{ha2016hypernetworks, wang2017learning}. The lifted structure loss~\cite{oh2016deep} introduces margins between many training instances.  The range loss~\cite{zhang2017range} enforces data in the same class to be close and those in different classes to be far apart.
The focal loss~\cite{lin2017focal} induces an online version of hard negative mining.
MetaModelNet~\cite{wang2017learning} learns a meta regression net from head classes and uses it to construct the classifier for tail classes.

Our dynamic meta-embedding combines the strengths of both metric learning and meta learning. On one hand, our direct feature is updated to ensure centroids for different classes are far from each other; On the other hand, our memory feature is generated on-the-fly in a meta learning fashion to effectively transfer knowledge to tail classes.

\vspace{2pt}
\noindent
\textbf{Few-Shot Learning.}
It is often formulated as meta learning~\cite{schmidhuber1993neural, bertinetto2016learning, ravi2016optimization, santoro2016meta, finn2017model, yang2018learning}. 
Matching Network~\cite{vinyals2016matching} learns a transferable feature matching metric to go beyond given classes. Prototypical Network~\cite{snell2017prototypical} maintains a set of separable class templates. 
Feature hallucination~\cite{hariharan2017low} and augmentation~\cite{wang2018low} are also shown effective.
Since these methods focus on novel classes, they often suffer a moderate performance drop for head classes.  There are a few exceptions. The few-shot learning without forgetting~\cite{gidaris2018dynamic} and incremental few-shot learning~\cite{ren2018incremental} attempt to remedy this issue by leveraging the duality between features and classifiers' weights~\cite{qiao2018few, qi2018low}. However, the training set used in all of these methods are balanced.  

In comparison, our OLTR learns from a more natural long-tailed training set.
Nevertheless, our work is closely related to meta learning with fast weight and associative memory~\cite{hinton1987using, schmidhuber1992learning, vinyals2016matching, duan2016rl, ba2016using, munkhdalai2017meta} to enable rapid adaptation. 
Compared to these prior arts, our memory feature has two advantages: {\bf 1)} It transfers knowledge to both head and tail classes adaptively via a learned concept selector; {\bf 2)} It is fully integrated into the network without episodic training, and is thus especially suitable for large-scale applications.  

\vspace{2pt}
\noindent
\textbf{Open-Set Recognition.}
Open-set recognition~\cite{scheirer2013toward, bendale2016towards}, or out-of-distribution detection~\cite{devries2018learning, liang2017enhancing}, aims to re-calibrate the sample confidence in the presence of open classes. 
One of the representative techniques is OpenMax~\cite{bendale2016towards}, which fits a Weibull distribution to the classifier's output logits.
However, when there are both open and tail classes, the distribution fitting could confuse the two.

Instead of calibrating the output logits, our OLTR approach incorporates the confidence estimation into feature learning and dynamically re-scale the meta-embedding \wrt to the learned visual memory.


\section{Our OLTR Model}

We propose to map an image to a feature space such that visual concepts can easily relate to each other based on a learned metric that respects the closed-world classification while acknowledging the novelty of the open world. Our model has two main modules (Fig.\ref{fig:pipeline}): \emph{dynamic meta-embedding} and \emph{modulated attention}.  The former relates and transfers knowledge between head and tail classes and the latter maintains discrimination between them.





\subsection{Dynamic Meta-Embedding}
Our dynamic meta-embedding combines a direct image feature and an associated memory feature, with the feature norm indicating the familiarity to known classes.

Consider a convolutional neural network (CNN) with a softmax output layer for classification. The  second-to-the-last layer can be viewed as the feature and  the last layer a linear classifier (cf.\ $\phi(\cdot)$ in Fig.~\ref{fig:pipeline}).
The feature and the classifier are jointly trained from big data in an end-to-end fashion.  Let $v^{direct}$ denote the {\it direct feature} extracted from an input image.  The final classification accuracy largely depends on the quality of this direct feature.

While a feed-forward CNN classifier works well with big training data~\cite{deng2009imagenet, krizhevsky2012imagenet}, it lacks sufficient supervised updates from small data in our tail classes.  We propose to enrich direct feature $v^{direct}$ with a memory feature $v^{memory}$ that relates visual concepts in a memory module.  This mechanism is similar to the memory popular in meta learning~\cite{santoro2016meta, munkhdalai2017meta}. We denote the resulting feature  {\it meta embedding} $v^{meta}$, and it is fed to the last layer for classification. 
Both our memory feature $v^{memory}$ and  meta-embedding $v^{meta}$ depend on direct feature $v^{direct}$.



Unlike the direct feature, the memory feature captures visual concepts from training classes, retrieved from a memory with a much shallower model.

\vspace{2pt}
\noindent
\textbf{Learning Visual Memory $M$.}
 We follow~\cite{hsu2017learning} on class structure analysis and adopt discriminative centroids as the basic building block.  Let $M$ denote the visual memory of all the training data, $M=\{c_{i}\}_{i=1}^{K}$ where $K$ is the number of training classes.
Compared to alternatives ~\cite{wen2016discriminative, snell2017prototypical}, this memory is appealing for our OLTR task:  It is almost effortlessly and jointly learned alongside the direct features $\{v_{n}^{direct}\}$, and it considers both intra-class compactness and inter-class discriminativeness.  

We compute centroids in two steps.
{\bf 1)} Neighborhood Sampling: We sample both intra-class and inter-class examples to compose a mini-batch during training. These examples are grouped by their class labels and the centroid $c_i$ of each group is updated by the direct feature of this mini-batch. 
{\bf 2)} Propagation: We alternatively update the direct feature $v^{direct}$ and the centroids to minimize the distance between each direct feature and the centroid of its group and maximize the distance to other centroids.  

\vspace{2pt}
\noindent
\textbf{Composing Memory Feature $v^{memory}$.}
For an input image, $v^{memory}$ shall enhance its direct feature when there is not enough training data (as in the tail class) to learn it well. The memory feature relates the centroids in the memory, transferring knowledge to the tail class:
\begin{align}
v^{memory} = o^T M := \sum_{i=1}^K o_{i}c_i, 
\end{align}
where $o\in\mathbb{R}^K$ is the coefficients hallucinated from the direct feature. We use a lightweight neural network to obtain the coefficients from the direct feature, $o = T_{hal}(v^{direct})$.



\begin{figure}[t]
  \centering
  \includegraphics[width=0.48\textwidth]{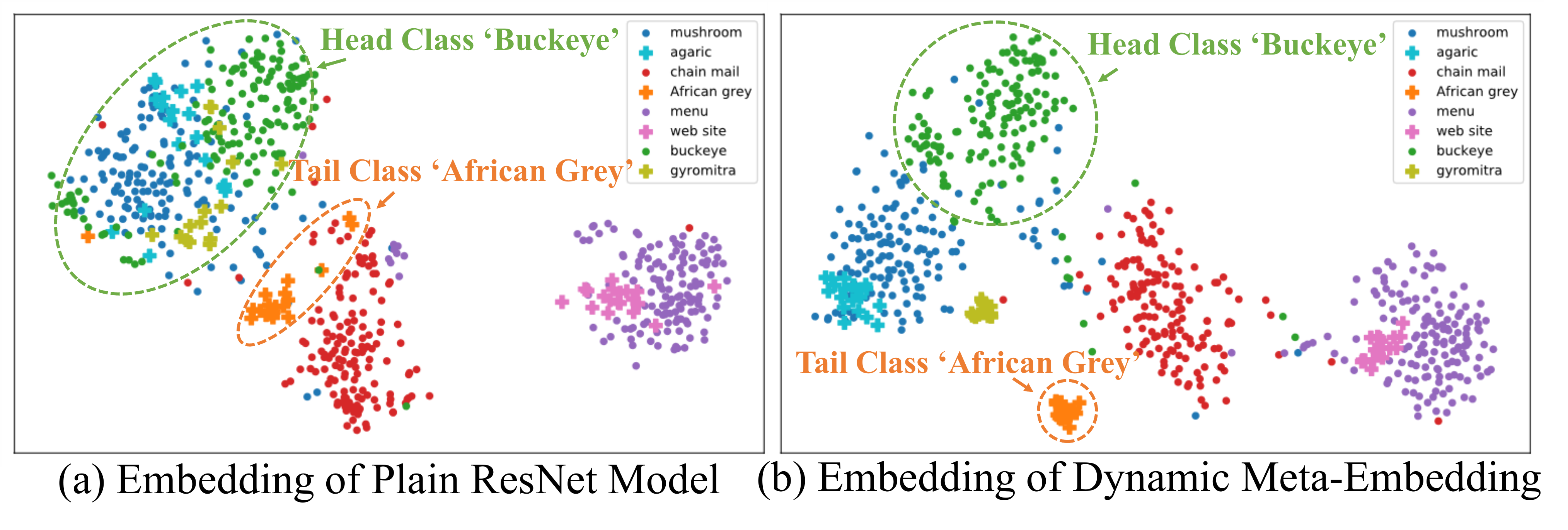}
  \caption{\textbf{t-SNE feature visualization} of (a) plain ResNet model (b) our \emph{dynamic meta-embedding}. Ours is more compact for both head and tail classes. }
  \label{fig:tsne}
\end{figure}

\vspace{2pt}
\noindent
{\bf Obtaining Dynamic Meta-Embedding}.
 $v^{meta}$  combines the direct feature and the memory feature, and is fed to the classifier for the final class prediction (Fig.~\ref{fig:tsne}):
\begin{equation}
v^{meta} = (1 / \gamma) \cdot (v^{direct} + e \otimes v^{memory}), \label{eq:embedding}
\end{equation}
where $\otimes$ denotes element-wise multiplication.
$\gamma>0$ is seemingly a redundant scalar for the closed-world classification tasks.  However, in the OLTR setting, it plays an important role in differentiating the examples of the training classes from those of the open-set.
$\gamma$ measures the reachability~\cite{savinov2018episodic} of an input's direct feature $v^{direct}$ to the memory $M$ --- the minimum distance between the direct feature and the discriminative centroids:
\begin{align}
\hspace{-3mm}\gamma := \text{reachability}(v^{direct}, M)=\min_i \;\|v^{direct} - c_{i}\|_{2}.
\end{align}
 When $\gamma $ is small, the  input likely belongs to a training class from which the centroids are derived, and a large reachability weight $1/\gamma$ is assigned to the resulting meta-embedding $v^{meta}$. Otherwise, the embedding is scaled down to an almost all-zero vector at the extreme.  Such a property is useful for encoding open classes.


We now describe the concept selector $e$ in Eq.~(\ref{eq:embedding}).  The direct feature is often good enough for the data-rich head classes, whereas the memory feature is more important for the data-poor tail classes.  To adaptively select them in a soft manner, we learn a lightweight network $T_{sel}(\cdot)$ with a $\tanh(\cdot)$ activation function:
\begin{equation}
e = \tanh(T_{sel}(v^{direct})).
\end{equation}


\subsection{Modulated Attention}
While {dynamic meta-embedding} facilitates feature sharing between head and tail classes, it is also vital to discriminate between them.  The  direct feature $v^{direct}$, e.g., the activation at the second-to-the-last layer in ResNet~\cite{he2016deep},
is able to fulfill this requirement to some extent. However, we find it beneficial to further enhance it with spatial attention, since discriminative cues of head and tail classes seem to be distributed at different locations in the image. 

\begin{figure}[t]
  \centering
  \includegraphics[width=0.48\textwidth]{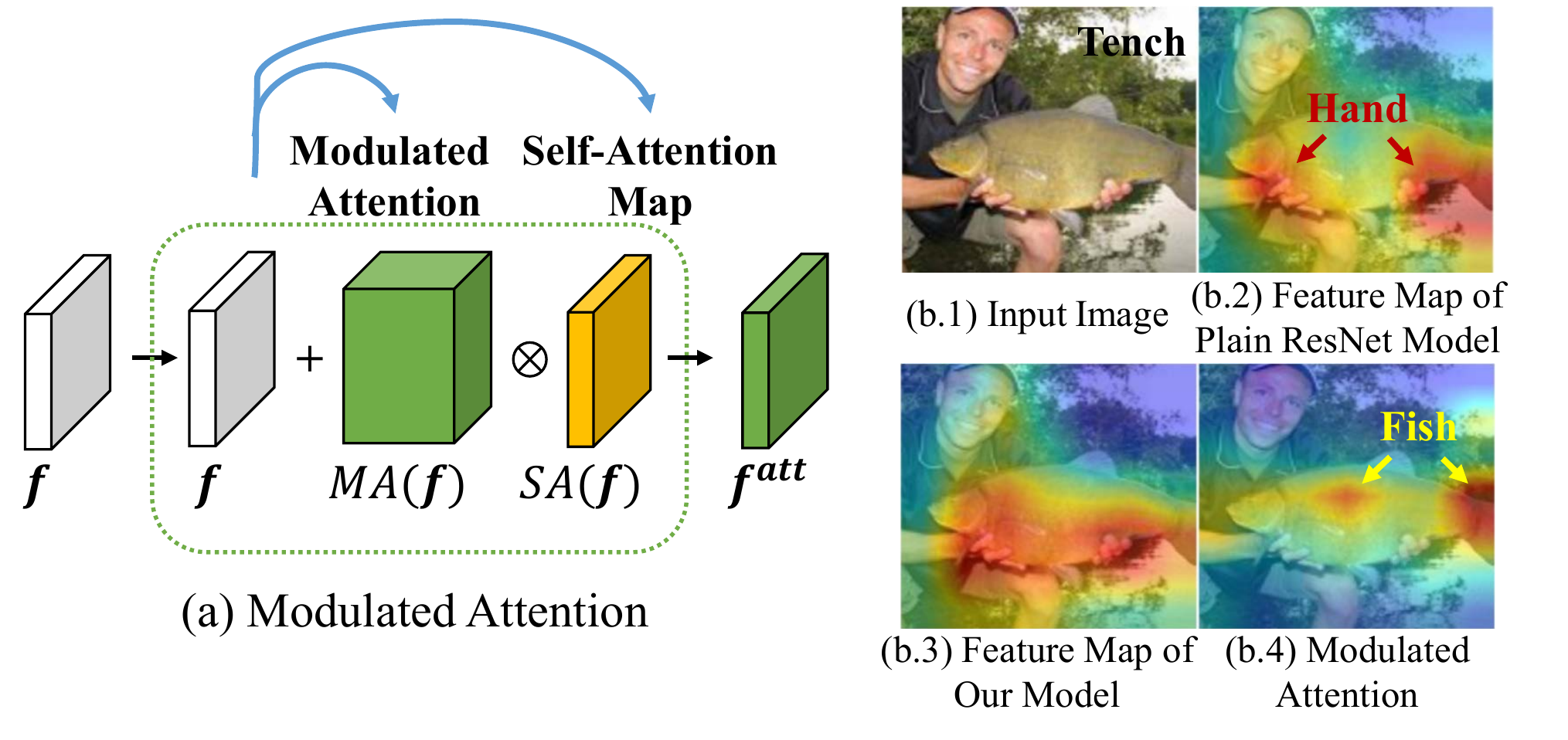}
  \caption{\textbf{Modulated attention} is spatial attention applied on self-attention maps (``attention on attention''). It encourages different classes to use different contexts, which helps maintain the discrimination between head and tail classes.}
  \label{fig:attention}
\end{figure}

Specifically, we propose \emph{modulated attention} to encourage samples of different classes to use different contexts.
Firstly, we compute a self-attention map $SA(f)$ from the input feature map by self-correlation~\cite{wang2017non}.
It is used as contextual information and added back (through skip connections) to the original feature map. 
The {modulated attention} $MA(f)$ is then designed as  conditional spatial attention applied to the self-attention map: $MA(f) \otimes SA(f)$, which allows examples to select different spatial contexts (Fig.~\ref{fig:attention}).  The final attention feature map becomes:
\begin{equation}
f^{att} = f + MA(f) \otimes SA(f),
\end{equation}
where $f$ is a feature map in CNN, $SA(\cdot)$ is the self-attention operation, and $MA(\cdot)$ is a conditional attention function~\cite{vaswani2017attention} with a softmax normalization.
 Sec.~\ref{sec:ablation} shows empirically  that our attention design achieves superior performance than the common practice of applying spatial attention to the input feature map. This modulated attention (Fig.~\ref{fig:attention}b) could be plugged into any feature layer of a CNN. Here, we modify the last feature map only.

\begin{figure*}[t]
\footnotesize
\parbox{.72\linewidth}{
\centering
\includegraphics[width=0.72\textwidth]{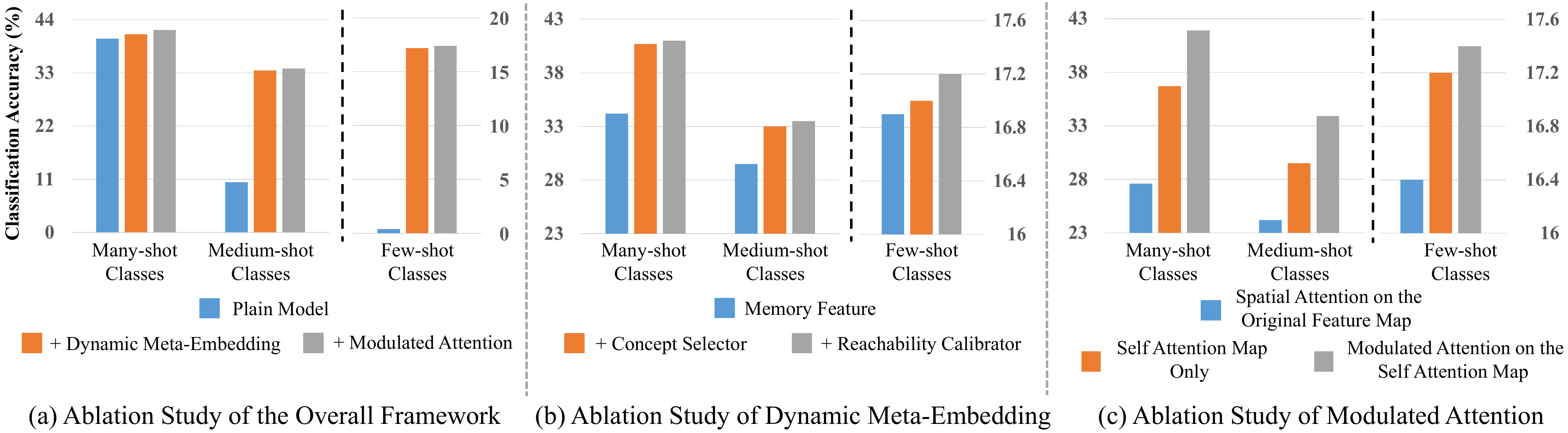}
\caption{\textbf{Results of ablation study.} Dynamic meta-embedding contributes most on medium-shot and few-shot classes while modulated attention helps maintain the discrimination of many-shot classes. (The performance is reported with \emph{open-set} top-$1$ classification accuracy on ImageNet-LT.)}
\label{fig:ablation}
}
\hfill
\parbox{.25\linewidth}{
\centering
\begin{tabular}{lc}
\Xhline{1pt}
\textbf{Method} & \textbf{Error (\%)} \\
\hline\hline
Softmax Pred.~\cite{hendrycks2016baseline} & 43.6 \\
Ours & 29.9 \\
ODIN~\cite{liang2017enhancing}$^{\dagger}$ & 24.6 \\
Ours$^{\dagger}$ & \textbf{18.0} \\
\Xhline{1pt}
\end{tabular}
\captionsetup{type=table} 
\caption{\textbf{Open class detection error (\%) comparison.} It is performed on the standard open-set benchmark, CIFAR100 + TinyImageNet (resized). ``$\dagger$'' denotes the setting where open samples are used to tune algorithmic parameters.}
\label{tab:openset}
}
\end{figure*}

\subsection{Learning}

\noindent
\textbf{Cosine Classifier.}
We adopt the cosine classifier~\cite{qi2018low,gidaris2018dynamic} to produce the final classification results.
Specifically, we normalize the meta-embeddings $\{v_{n}^{meta}\}$, where $n$ stands for the $n$-th input as well as the weight vectors $\{w_i\}_{i=1}^K$ of the classifier $\phi(\cdot)$ (no bias term):
\begin{equation}
\begin{split}
v_{n}^{meta} &= \frac{\|v_{n}^{meta}\|^{2}}{1+\|v_{n}^{meta}\|^{2}} \cdot \frac{v_{n}^{meta}}{\|v_{n}^{meta}\|}, \\
w_{k} &= \frac{w_{k}}{\|w_{k}\|}.
\end{split}
\end{equation}
The normalization strategy for the meta-embedding is a non-linear squashing function \cite{sabour2017dynamic}
which ensures that vectors of small magnitude are shrunk to almost zeros while vectors of big magnitude are normalized to the length slightly below $1$. This function helps amplify the effect of the reachability $\gamma$ (cf.\ Eq.~(\ref{eq:embedding})).

\vspace{2pt}
\noindent
\textbf{Loss Function.}
Since all our modules are differentiable, our model can be trained end-to-end by alternatively updating the  centroids $\{c_{i}\}_{i=1}^{K}$ and the \emph{dynamic meta-embedding} $v_{n}^{meta}$.
The final loss function $L$ is a combination of the cross-entropy classification loss $L_{CE}$ and the large-margin loss between the embeddings and the centroids $L_{LM}$:
\begin{equation}
L = \sum_{n=1}^{N} L_{CE}(v_{n}^{meta}, y_{n}) + \lambda \cdot L_{LM}(v_{n}^{meta}, \{c_{i}\}_{i=1}^{K}),
\end{equation}
where $\lambda$ is set to $0.1$ in our experiments 
via observing the accuracy curve on validation set.



\section{Experiments}

\noindent
\textbf{Datasets.}
%
We curate three open long-tailed benchmarks, ImageNet-LT (object-centric), Places-LT (scene-centric), and MS1M-LT (face-centric), respectively.
\begin{enumerate}[leftmargin=*]
\item ImageNet-LT: We construct a long-tailed version of the original ImageNet-2012~\cite{deng2009imagenet} by sampling a subset  following the Pareto distribution with the power value $\alpha$=6. Overall, it has $115.8$K images from $1000$ categories, with  maximally $1280$ images per class and minimally $5$ images per class. 
The additional classes of images in ImageNet-2010 are used as the open set.
We make the test set balanced.
\item Places-LT: A long-tailed version of Places-2~\cite{zhou2018places} is constructed in a similar way. It contains $184.5$K images from $365$ categories, with the maximum of $4980$ images per class and the minimum of $5$ images per class. 
The gap between the head and tail classes are even larger than ImageNet-LT. We use the test images from Places-Extra69 as the additional open-set.
\item MS1M-LT: 
To create a long-tailed version of the MS1M-ArcFace dataset~\cite{guo2016ms, deng2018arcface}, we sample images for each identity with a probability proportional to the image numbers of each identity. It results in $887.5$K images and $74.5$K identities, with a long-tailed distribution.
To inspect the generalization ability of our approach, the performance is evaluated on the MegaFace benchmark~\cite{kemelmacher2016megaface}, which has no identity overlap with MS1M-ArcFace.
\end{enumerate}

\noindent
\textbf{Network Architectures.}
Following~\cite{hariharan2017low, wang2018low, gidaris2018dynamic}, we employ the scratch ResNet-10~\cite{he2016deep} as our backbone network for ImageNet-LT.
To make a fair comparison with~\cite{wang2017learning}, the pre-trained ResNet-152~\cite{he2016deep} is used as the backbone network for Places-LT.
For MS1M-LT, the popular pre-trained ResNet-50~\cite{he2016deep} is the backbone network.

\vspace{2pt}
\noindent
\textbf{Evaluation Metrics.}
We evaluate the performance of each method under both  the \emph{closed-set} (test set contains no unknown classes) and \emph{open-set} (test set contains unknown classes) settings to highlight their differences.
Under each setting, besides the overall top-$1$ classification accuracy~\cite{gidaris2018dynamic} over all classes, we also calculate the accuracy of three disjoint subsets: \emph{many-shot classes} (classes each with over training 100 samples), \emph{medium-shot classes} (classes each with 20$\sim$100 training samples) and \emph{few-shot classes} (classes under 20 training samples).
This helps us understand the detailed characteristics of each method.
For the \emph{open-set} setting, the \emph{F-measure} is also reported for a balanced treatment of precision and recall following~\cite{bendale2016towards}.
For determining open classes, the $softmax$ probability threshold is initially set as $0.1$, while a more detailed analysis is provided in Sec.~\ref{sec:analysis}.


\vspace{2pt}
\noindent
\textbf{Competing Methods.}
We choose for comparison state-of-the-art methods from different fields dealing with the open long-tailed data, including: (1) \emph{metric learning}: Lifted Loss~\cite{oh2016deep}, (2) \emph{hard negative mining}: Focal Loss~\cite{lin2017focal}, (3) \emph{feature regularization}: Range Loss~\cite{zhang2017range}, (4) \emph{few-shot learning}: FSLwF~\cite{gidaris2018dynamic}, (5) \emph{long-tailed modeling}: MetaModelNet~\cite{wang2017learning}, and (6) \emph{open-set detection}: Open Max~\cite{bendale2016towards}. 
We apply these methods on the same backbone networks as ours for a fair comparison.
We also enable them with class-aware mini-batch sampling~\cite{shen2016relay} for effective learning.
Since Model Regression~\cite{wang2016learning} and MetaModelNet~\cite{wang2017learning} are the most related to our work, we directly contrast our results to the numbers reported in their paper.

\begin{table*}
\footnotesize

\begin{subtable}{\linewidth}
\centering
\bgroup
\begin{tabular}{l|cccc|cccc}
\Xhline{1pt}
\textbf{Backbone Net} & \multicolumn{4}{c|}{\footnotesize{\textbf{closed-set setting}}} & \multicolumn{4}{c}{\footnotesize{\textbf{open-set setting}}} \\
ResNet-10 & $>100$ & $\leqslant100$ \& $>20$ & $<20$ && $>100$ & $\leqslant100$ \& $>20$ & $<20$ &\\
\textbf{Methods} & \textbf{Many-shot} & \textbf{Medium-shot} & \textbf{Few-shot} & \textbf{Overall} & \textbf{Many-shot} & \textbf{Medium-shot} & \textbf{Few-shot} & \textbf{F-measure} \\
\hline\hline
Plain Model~\cite{he2016deep} & 40.9 & 10.7 & 0.4 & 20.9 & 40.1 & 10.4 & 0.4 & 0.295 \\
Lifted Loss~\cite{oh2016deep} & 35.8 & 30.4 & 17.9 & 30.8 & 34.8 & 29.3 & 17.4 & 0.374 \\
Focal Loss~\cite{lin2017focal} & 36.4 & 29.9 & 16 & 30.5 & 35.7 & 29.3 & 15.6 & 0.371 \\
Range Loss~\cite{zhang2017range} & 35.8 & 30.3 & 17.6 & 30.7 & 34.7 & 29.4 & 17.2 & 0.373 \\
~~~+~OpenMax~\cite{bendale2016towards} & - & - & - & - & 35.8 & 30.3 & \textbf{17.6} & 0.368 \\
FSLwF~\cite{gidaris2018dynamic} & 40.9 & 22.1 & 15 & 28.4 & 40.8 & 21.7 & 14.5 & 0.347 \\
\hline
Ours & \textbf{43.2} & \textbf{35.1} & \textbf{18.5} & \textbf{35.6} & \textbf{41.9} & \textbf{33.9} & 17.4 & \textbf{0.474} \\
\Xhline{1pt}
\end{tabular}\vspace{-5pt}
\egroup
\subcaption{ Top-$1$ classification accuracy on ImageNet-LT.}
\end{subtable}

\begin{subtable}{\linewidth}
\centering
\bgroup
\begin{tabular}{l|cccc|cccc}
\Xhline{1pt}
\textbf{Backbone Net} & \multicolumn{4}{c|}{\footnotesize{\textbf{closed-set setting}}} & \multicolumn{4}{c}{\footnotesize{\textbf{open-set setting}}} \\
ResNet-152 & $>100$ & $\leqslant100$ \& $>20$ & $<20$ && $>100$ & $\leqslant100$ \& $>20$ & $<20$ &\\
\textbf{Methods} & \textbf{Many-shot} & \textbf{Medium-shot} & \textbf{Few-shot} & \textbf{Overall} & \textbf{Many-shot} & \textbf{Medium-shot} & \textbf{Few-shot} & \textbf{F-measure} \\
\hline\hline
Plain Model~\cite{he2016deep} & \textbf{45.9} & 22.4 & 0.36 & 27.2 & \textbf{45.9} & 22.4 & 0.36 & 0.366 \\
Lifted Loss~\cite{oh2016deep} & 41.1 & 35.4 & 24 & 35.2 & 41 & 35.2 & 23.8 & 0.459 \\
Focal Loss~\cite{lin2017focal} & 41.1 & 34.8 & 22.4 & 34.6 & 41 & 34.8 & 22.3 & 0.453 \\
Range Loss~\cite{zhang2017range} & 41.1 & 35.4 & 23.2 & 35.1 & 41 & 35.3 & 23.1 & 0.457 \\
~~~+~OpenMax~\cite{bendale2016towards} & - & - & - & - & 41.1 & 35.4 & 23.2 & 0.458 \\
FSLwF~\cite{gidaris2018dynamic} & 43.9 & 29.9 & \textbf{29.5} & 34.9 & 38.1 & 19.5 & 14.8 & 0.375 \\
\hline
Ours & 44.7 & \textbf{37} & 25.3 & \textbf{35.9} & 44.6 & \textbf{36.8} & \textbf{25.2} & \textbf{0.464} \\
\Xhline{1pt}
\end{tabular}\vspace{-5pt}
\egroup
\subcaption{ Top-$1$ classification accuracy on Places-LT.}
\end{subtable}
\caption{\textbf{Benchmarking results on (a) ImageNet-LT and (b) Places-LT.} Our approach provides a comprehensive treatment to all the many/medium/few-shot classes as well as the open classes, achieving substantial advantages on all aspects.}
\label{tab:benchmark_imagenet}
\end{table*}


\begin{table*}
\footnotesize
\parbox{.7\linewidth}{
\centering
\begin{tabular}{l|ccccc|cc}
\Xhline{1pt}
\textbf{Backbone Net} & \multicolumn{7}{c}{\footnotesize{\textbf{MegaFace Identification Rate}}}  \\
ResNet-50 & $\geqslant5$ & $<5$ \& $\geqslant2$ & $<2$ \& $\geqslant1$ & $=0$ & & \multicolumn{2}{c}{Sub-Groups} \\
\textbf{Methods} & \textbf{Many-shot} & \textbf{Few-shot} & \textbf{One-shot} & \textbf{Zero-shot} & \textbf{Full Test} & \textbf{Male} & \textbf{Female} \\
\hline\hline
Plain Model~\cite{he2016deep} & 80.64 & 71.98 & 84.60 & 77.72 & 73.88 & 78.30 & 78.70 \\ 
Range Loss~\cite{zhang2017range} & 78.60 & 71.36 & 83.14 & 77.40 & 72.17 & - & - \\ 
Ours & \textbf{80.82} & \textbf{72.44} & \textbf{87.60} & \textbf{79.50} & \textbf{74.51} & \textbf{79.04} & \textbf{79.08} \\
\Xhline{1pt}
\end{tabular}
}
\hfill
\parbox{.2\linewidth}{
\centering
\begin{tabular}{lc}
\Xhline{1pt}
\textbf{Method} & \textbf{Acc.} \\
\hline\hline
Plain Model~\cite{he2016deep} & 48.0 \\
Cost-Sensitive~\cite{huang2016learning} & 52.4 \\
Model Reg.~\cite{wang2016learning} & 54.7 \\
MetaModelNet~\cite{wang2017learning} & 57.3 \\
\hline
Ours & \textbf{58.7} \\
\Xhline{1pt}
\end{tabular}
}
\caption{
\textbf{Benchmarking results on MegaFace (\emph{left}) and SUN-LT (\emph{right}).} Our approach achieves the best performance on natural-world datasets when compared to other state-of-the-art methods. Furthermore, our approach achieves across-board improvements on both `male' and `female' sub-groups.}
\label{tab:benchmark_megaface}
\end{table*}

\subsection{Ablation Study}
\label{sec:ablation}

We firstly investigate the merit of each module in our framework.
The performance is reported with \emph{open-set} top-$1$ classification accuracy on ImageNet-LT.

\vspace{2pt}
\noindent
\textbf{Effectiveness of the Dynamic Meta-Embedding.}
Recall that the dynamic meta-embedding consists of three main components: memory feature, concept selector, and confidence calibrator.
From Fig.~\ref{fig:ablation} (b), we observe that the combination of the memory feature and concept selector leads to large improvements on all three shots.
It is because the obtained memory feature transfers useful visual concepts among classes.
Another observation is that the confidence calibrator is the most effective on few-shot classes.
The reachability estimation inside the confidence calibrator helps distinguish tail classes from open classes.

\vspace{2pt}
\noindent
\textbf{Effectiveness of the Modulated Attention.}
We observe from Fig.~\ref{fig:ablation} (a) that, compared to medium-shot classes, the modulated attention contributes more to the discrimination between many-shot and few-shot classes. Fig.~\ref{fig:ablation} (c) further validates that the modulated attention is more effective than directly applying spatial attention on feature maps.
It implies that adaptive contexts selection is easier to learn than the conventional feature selection.

\begin{figure}[t]
  \centering
  \includegraphics[width=0.49\textwidth]{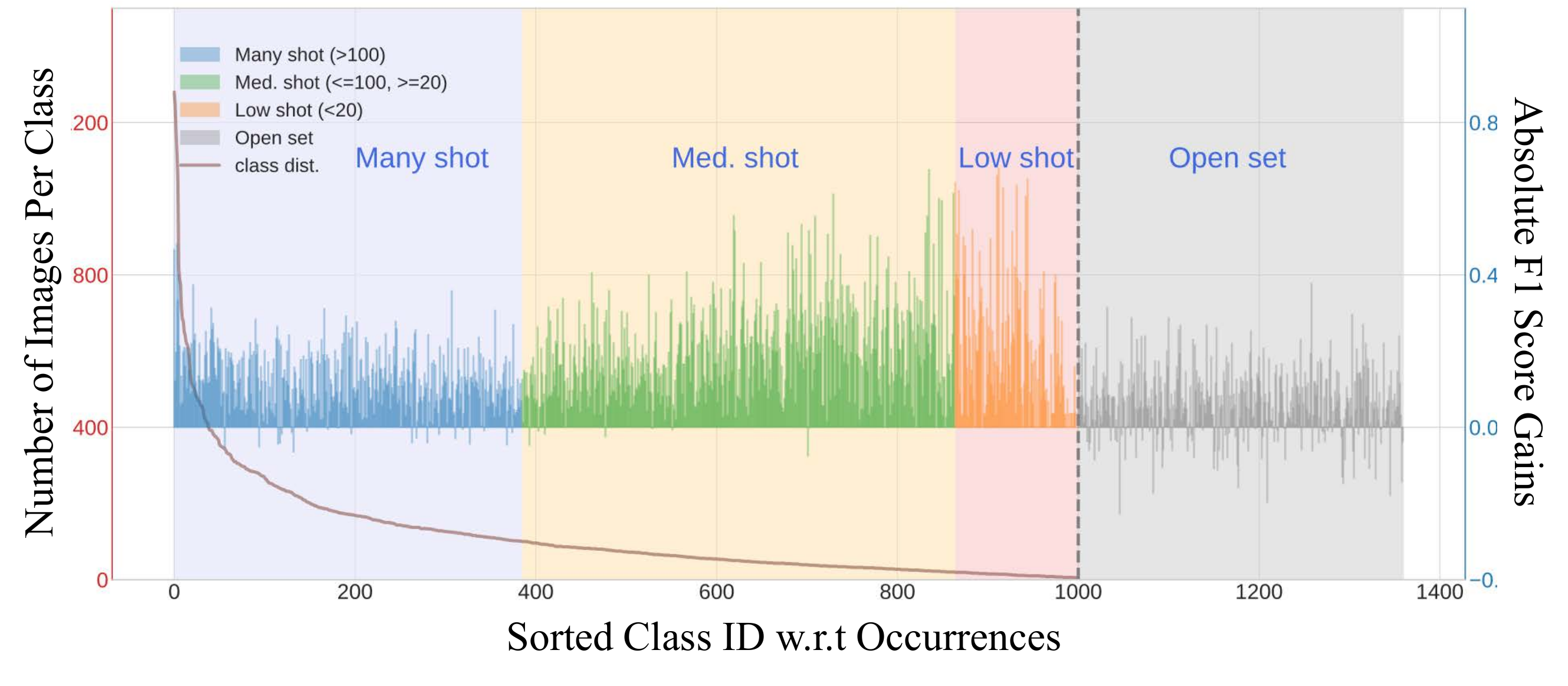}
  \caption{\textbf{The absolute F1 score of our method over the plain model.} Ours has across-the-board performance gains \wrt many/medium/few-shot and open classes.}
  \label{fig:acc_gain}
\end{figure}

\vspace{2pt}
\noindent
\textbf{Effectiveness of the Reachability Calibration.}
To further demonstrate the merit of reachability calibration for open-world setting, we conduct additional experiments following the standard settings in~\cite{hendrycks2016baseline, liang2017enhancing} (CIFAR100 + TinyImageNet(resized)). The results are listed in Table~\ref{tab:openset}, where our approach shows favorable performance over standard open-set methods~\cite{hendrycks2016baseline, liang2017enhancing}.

\begin{figure*}[t]
  \centering
  \includegraphics[width=1.0\textwidth]{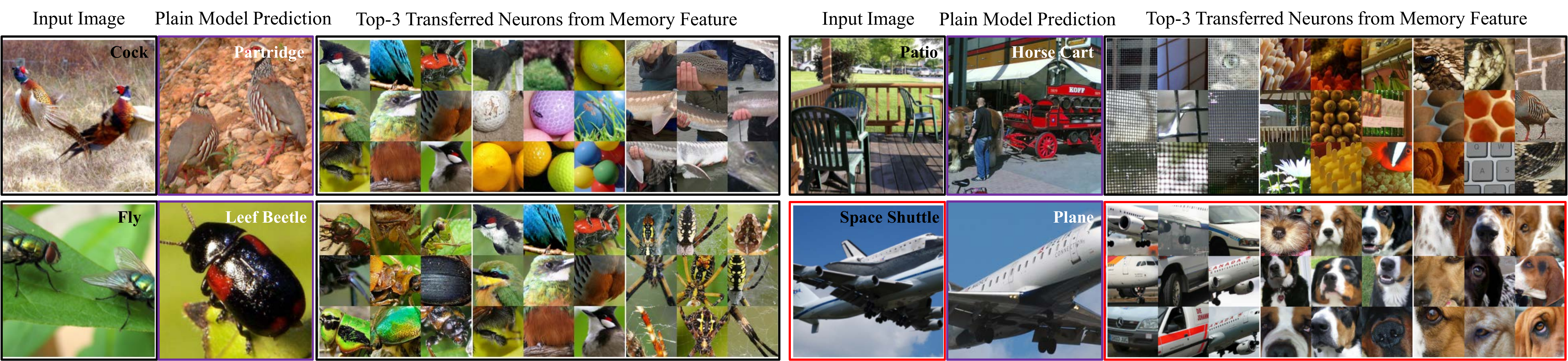}
  \caption{\textbf{Examples of the top-$3$ infused visual concepts from memory feature.} Except for the bottom right failure case 
  (marked in red), all the other three input images are misclassified by the plain model and correctly classified by our model. For example, to classify the top left image which belongs to a tail class `cock', our approach has learned to transfer visual concepts that represents ``bird head'', ``round shape'' and ``dotted texture'' respectively. }
  \label{fig:neuron}
\end{figure*}

\subsection{Result Comparisons}

We extensively evaluate the performance of various representative methods on our benchmarks.

\vspace{2pt}
\noindent
\textbf{ImageNet-LT.}
Table~\ref{tab:benchmark_imagenet} (a) shows the performance comparison of different methods.
We have the following observations.
Firstly, both Lifted Loss~\cite{oh2016deep} and Focal Loss~\cite{lin2017focal} greatly boost the performance of few-shot classes by enforcing feature regularization.
However, they also sacrifice the performance on many-shot classes  since there are no built-in mechanism of adaptively handling  samples of different shots. 
Secondly, OpenMax~\cite{bendale2016towards} improves the results under the open-set setting.
However, the accuracy degrades when it is evaluated with \emph{F-measure}, which considers both precision and recall in open-set.
When the open classes are compounded with the tail classes, it becomes challenging to perform the distribution fitting that~\cite{bendale2016towards} requires.
Lastly, though the few-shot learning without forgetting approach~\cite{gidaris2018dynamic} retains the many-shot class accuracy, it has difficulty dealing with the imbalanced base classes which are lacked in the current few-shot paradigm.
As demonstrated in Fig.~\ref{fig:acc_gain}, our approach provides a comprehensive treatment to all the many/medium/few-shot classes as well as the open classes,  achieving substantial improvements on all aspects.

\vspace{2pt}
\noindent
\textbf{Places-LT.}
Similar observations can be made on the Places-LT benchmark as shown in Table~\ref{tab:benchmark_imagenet} (b).
With a much stronger baseline (\ie pre-trained ResNet-152), our approach still consistently outperforms other alternatives under both the closed-set and open-set settings.
The advantage is even more profound under the \emph{F-measure}.

\vspace{2pt}
\noindent
\textbf{MS1M-LT.}
We train on the MS1M-LT dataset and report results on the MegaFace identification track, which is a standard benchmark in the face recognition field.
Since the face identities in the training set and the test set are disjoint, we adopt an indirect way to partition the testing set into the subsets of different shots. We approximate the pseudo shots of each test sample by counting the number of training samples that are similar to it by at least a threshold (feature similarity greater than $0.7$). Apart from many-shot, few-shot, one-shot subsets, we also obtain a zero-shot subset, for which we cannot find any sufficiently similar samples in the training set.
It can be observed that our approach has the most advantage on one-shot identities ($3.0\%$ gains) and zero-shot identities ($1.8\%$ gains) as shown in Table~\ref{tab:benchmark_megaface} (\emph{left}).

\vspace{2pt}
\noindent
\textbf{SUN-LT.}
To directly compare with~\cite{wang2016learning} and~\cite{wang2017learning}, we also test on the SUN-LT benchmark they provided.
The final results are listed in Table~\ref{tab:benchmark_megaface} (\emph{right}).
Instead of learning a series of classifier transformations, our approach transfers visual knowledge among features and achieves a $1.4\%$ improvement over the prior best.
Note that our approach also incurs much less computational cost since MetaModelNet~\cite{wang2017learning} requires a recursive training procedure.

\vspace{2pt}
\noindent
\textbf{Indication for Fairness.}
Here we report the sensitive attribute performance on MS1M-LT.
The last two columns in Table~\ref{tab:benchmark_megaface} show that our approach achieves across-board improvements on both `male' and `female' sub-groups, which has an implication for effective fairness learning.

\begin{figure}
  \centering
  \includegraphics[width=0.48\textwidth]{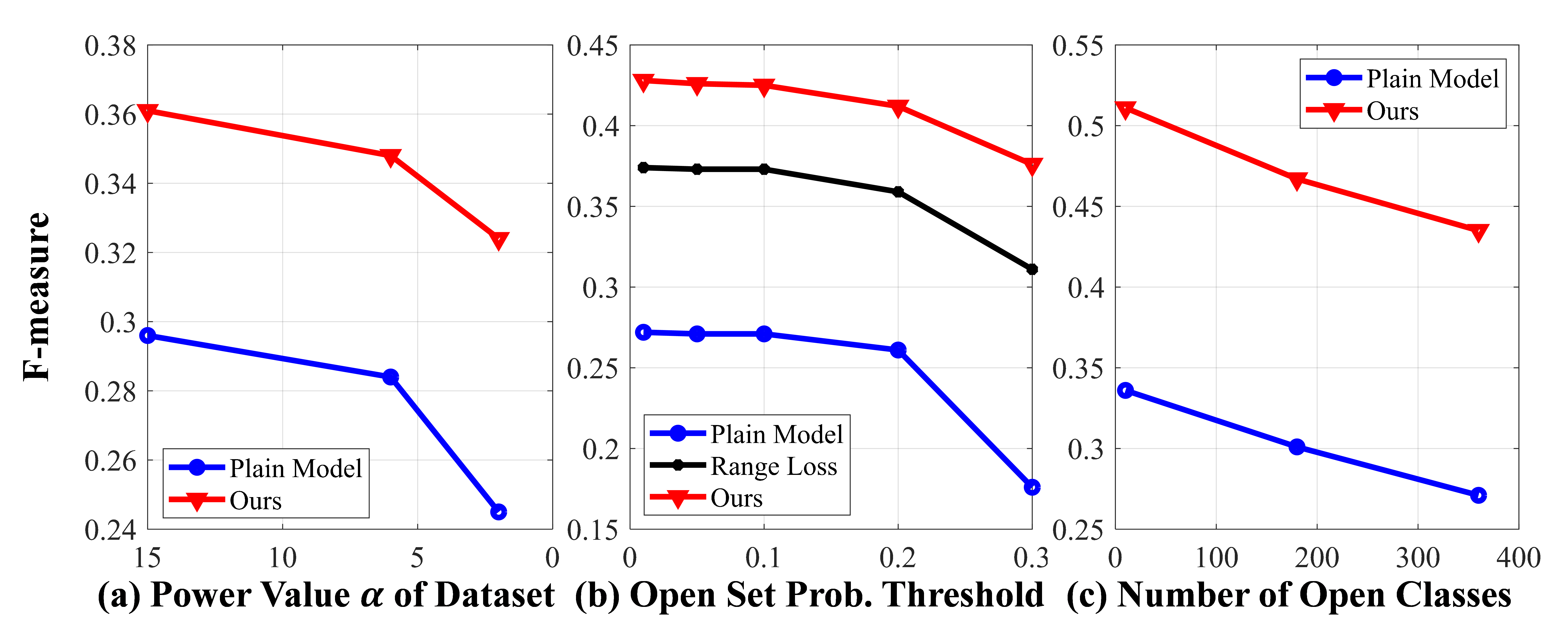}
  \caption{\textbf{The influence of (a) dataset longtail-ness, (b) open-set probability threshold, and (c) the number of open classes.} As the dataset becomes more imbalanced, our approach only undergoes a moderate performance drop. Our approach also demonstrates great robustness to the contamination of open classes.}
  \label{fig:analysis}
\end{figure}

\subsection{Further Analysis}
\label{sec:analysis}

Finally we visualize and analyze some influencing aspects in our framework as well as typical failure cases.

\vspace{2pt}
\noindent
\textbf{What memory feature has Infused.}
Here we inspect the visual concepts that memory feature has infused by visualizing its top activating neurons as shown in Fig.~\ref{fig:neuron}.
Specifically, for each input image, we identify its top-$3$ transferred neurons in memory feature.
And each neuron is visualized by a collection of highest activated patches~\cite{zeiler2014visualizing} over the whole training set. 
For example, to classify the top left image which belongs to a tail class `cock', our approach has learned to transfer visual concepts that represents ``bird head'', ``round shape'' and ``dotted texture'' respectively. 
After feature infusion, the dynamic meta-embedding becomes more informative and discriminative.

\vspace{2pt}
\noindent
\textbf{Influence of Dataset Longtail-ness.}
The longtail-ness of the dataset (\eg the degree of imbalance of the class distribution) could have an impact on the model performance. 
For faster investigating, here the weights of the backbone network are freezed during training.
From Fig.~\ref{fig:analysis} (a), we observe that as the dataset becomes more imbalanced (\ie power value $\alpha$ decreases), our approach only undergoes a moderate performance drop.
Dynamic meta-embedding enables effective knowledge transfer among data-abundant and data-scarce classes.

\vspace{2pt}
\noindent
\textbf{Influence of Open-Set Prob. Threshold.}
The performance change \wrt the open-set probability threshold is demonstrated in Fig.~\ref{fig:analysis} (b).
Compared to the plain model~\cite{he2016deep} and range loss~\cite{zhang2017range}, the performance of our approach changes steadily as the open-set threshold rises.
The reachability estimator in our framework helps calibrate the sample confidence, thus enhancing robustness to open classes.

\vspace{2pt}
\noindent
\textbf{Influence of the Number of Open Classes.}
Finally we investigate performance change \wrt the number of open classes.
Fig.~\ref{fig:analysis} (c) indicates that our approach demonstrates great robustness to the contamination of open classes.

\vspace{2pt}
\noindent
\textbf{Failure Cases.}
Since our approach encourages the feature infusion among classes, it slightly sacrifices the fine-grained discrimination for the promotion of under-representative classes. 
One typical failure case of our approach is the confusion between many-shot and medium-shot classes.
For example, the bottom right image in Fig.~\ref{fig:neuron} is misclassified into `airplane' because some cross-category traits like ``nose shape'' and ``eye shape'' are infused.
We plan to explore feature disentanglement~\cite{bengio2013representation} to alleviate this trade-off issue.

\section{Conclusions}

We introduce the OLTR task that learns from natural long-tail open-end distributed data and optimizes the overall accuracy over a balanced test set.
We propose an integrated OLTR algorithm, dynamic meta-embedding,
in order to share visual knowledge between head and tail classes and  to reduce confusion between tail and open classes.
We validate our method on three curated large-scale OLTR benchmarks (ImageNet-LT, Places-LT and MS1M-LT).  Our publicly available code and data would enable future research that is directly transferable to real-world applications.

\vspace{2pt}
\noindent
\textbf{Acknowledgements.}
{\small
This research was supported, in part, by SenseTime Group Limited, NSF IIS 1835539, Berkeley Deep Drive, DARPA, and US Government fund through Etegent Technologies on Low-Shot Detection in Remote Sensing Imagery.  The views, opinions and/or findings expressed are those of the author and should not be interpreted as representing the official views or policies of the Department of Defense or the U.S. Government.
}

{\small
\bibliographystyle{ieee_fullname}
\bibliography{reference}

\begin{thebibliography}{10}\itemsep=-1pt

\bibitem{anne2018women}
Lisa Anne~Hendricks, Kaylee Burns, Kate Saenko, Trevor Darrell, and Anna
  Rohrbach.
\newblock Women also snowboard: Overcoming bias in captioning models.
\newblock In {\em ECCV}, 2018.

\bibitem{ba2016using}
Jimmy Ba, Geoffrey~E Hinton, Volodymyr Mnih, Joel~Z Leibo, and Catalin Ionescu.
\newblock Using fast weights to attend to the recent past.
\newblock In {\em NIPS}, 2016.

\bibitem{bendale2016towards}
Abhijit Bendale and Terrance~E Boult.
\newblock Towards open set deep networks.
\newblock In {\em CVPR}, 2016.

\bibitem{bengio2015the}
Samy Bengio.
\newblock The battle against the long tail.
\newblock In {\em Talk on Workshop on Big Data and Statistical Machine
  Learning}, 2015.

\bibitem{bengio2013representation}
Yoshua Bengio, Aaron Courville, and Pascal Vincent.
\newblock Representation learning: A review and new perspectives.
\newblock {\em TPAMI}, 2013.

\bibitem{bertinetto2016learning}
Luca Bertinetto, Jo{\~a}o~F Henriques, Jack Valmadre, Philip Torr, and Andrea
  Vedaldi.
\newblock Learning feed-forward one-shot learners.
\newblock In {\em NIPS}, 2016.

\bibitem{cui2018large}
Yin Cui, Yang Song, Chen Sun, Andrew Howard, and Serge Belongie.
\newblock Large scale fine-grained categorization and domain-specific transfer
  learning.
\newblock In {\em CVPR}, 2018.

\bibitem{deng2009imagenet}
Jia Deng, Wei Dong, Richard Socher, Li-Jia Li, Kai Li, and Li Fei-Fei.
\newblock Imagenet: A large-scale hierarchical image database.
\newblock In {\em CVPR}, 2009.

\bibitem{deng2018arcface}
Jiankang Deng, Jia Guo, and Stefanos Zafeiriou.
\newblock Arcface: Additive angular margin loss for deep face recognition.
\newblock {\em arXiv preprint arXiv:1801.07698}, 2018.

\bibitem{devries2018learning}
Terrance DeVries and Graham~W Taylor.
\newblock Learning confidence for out-of-distribution detection in neural
  networks.
\newblock {\em arXiv preprint arXiv:1802.04865}, 2018.

\bibitem{dong2017class}
Qi Dong, Shaogang Gong, and Xiatian Zhu.
\newblock Class rectification hard mining for imbalanced deep learning.
\newblock In {\em ICCV}, 2017.

\bibitem{duan2016rl}
Yan Duan, John Schulman, Xi Chen, Peter~L Bartlett, Ilya Sutskever, and Pieter
  Abbeel.
\newblock Rl$^{2}$: Fast reinforcement learning via slow reinforcement
  learning.
\newblock {\em arXiv preprint arXiv:1611.02779}, 2016.

\bibitem{dwork2012fairness}
Cynthia Dwork, Moritz Hardt, Toniann Pitassi, Omer Reingold, and Richard Zemel.
\newblock Fairness through awareness.
\newblock In {\em The 3rd innovations in theoretical computer science
  conference}, 2012.

\bibitem{finn2017model}
Chelsea Finn, Pieter Abbeel, and Sergey Levine.
\newblock Model-agnostic meta-learning for fast adaptation of deep networks.
\newblock {\em arXiv preprint arXiv:1703.03400}, 2017.

\bibitem{gidaris2018dynamic}
Spyros Gidaris and Nikos Komodakis.
\newblock Dynamic few-shot visual learning without forgetting.
\newblock In {\em CVPR}, 2018.

\bibitem{guo2016ms}
Yandong Guo, Lei Zhang, Yuxiao Hu, Xiaodong He, and Jianfeng Gao.
\newblock Ms-celeb-1m: A dataset and benchmark for large-scale face
  recognition.
\newblock In {\em ECCV}, 2016.

\bibitem{ha2016hypernetworks}
David Ha, Andrew Dai, and Quoc~V Le.
\newblock Hypernetworks.
\newblock {\em arXiv preprint arXiv:1609.09106}, 2016.

\bibitem{hariharan2017low}
Bharath Hariharan and Ross~B Girshick.
\newblock Low-shot visual recognition by shrinking and hallucinating features.
\newblock In {\em ICCV}, 2017.

\bibitem{he2008learning}
Haibo He and Edwardo~A Garcia.
\newblock Learning from imbalanced data.
\newblock {\em TKDE}, 2008.

\bibitem{he2016deep}
Kaiming He, Xiangyu Zhang, Shaoqing Ren, and Jian Sun.
\newblock Deep residual learning for image recognition.
\newblock In {\em CVPR}, 2016.

\bibitem{hendrycks2016baseline}
Dan Hendrycks and Kevin Gimpel.
\newblock Baseline for detecting misclassified and out-of-distribution examples
  in neural networks.
\newblock In {\em ICLR}, 2017.

\bibitem{hinton1987using}
Geoffrey~E Hinton and David~C Plaut.
\newblock Using fast weights to deblur old memories.
\newblock In {\em Proceedings of the ninth annual conference of the Cognitive
  Science Society}, 1987.

\bibitem{hsu2017learning}
Yen-Chang Hsu, Zhaoyang Lv, and Zsolt Kira.
\newblock Learning to cluster in order to transfer across domains and tasks.
\newblock {\em arXiv preprint arXiv:1711.10125}, 2017.

\bibitem{huang2016learning}
Chen Huang, Yining Li, Chen~Change Loy, and Xiaoou Tang.
\newblock Learning deep representation for imbalanced classification.
\newblock In {\em CVPR}, 2016.

\bibitem{kemelmacher2016megaface}
Ira Kemelmacher-Shlizerman, Steven~M Seitz, Daniel Miller, and Evan Brossard.
\newblock The megaface benchmark: 1 million faces for recognition at scale.
\newblock In {\em CVPR}, 2016.

\bibitem{krizhevsky2012imagenet}
Alex Krizhevsky, Ilya Sutskever, and Geoffrey~E Hinton.
\newblock Imagenet classification with deep convolutional neural networks.
\newblock In {\em NIPS}, 2012.

\bibitem{lake2015human}
Brenden~M Lake, Ruslan Salakhutdinov, and Joshua~B Tenenbaum.
\newblock Human-level concept learning through probabilistic program induction.
\newblock {\em Science}, 2015.

\bibitem{liang2017enhancing}
Shiyu Liang, Yixuan Li, and R Srikant.
\newblock Enhancing the reliability of out-of-distribution image detection in
  neural networks.
\newblock In {\em ICLR}, 2018.

\bibitem{lin2017focal}
Tsung-Yi Lin, Priyal Goyal, Ross Girshick, Kaiming He, and Piotr Doll{\'a}r.
\newblock Focal loss for dense object detection.
\newblock In {\em ICCV}, 2017.

\bibitem{lin2014microsoft}
Tsung-Yi Lin, Michael Maire, Serge Belongie, James Hays, Pietro Perona, Deva
  Ramanan, Piotr Doll{\'a}r, and C~Lawrence Zitnick.
\newblock Microsoft coco: Common objects in context.
\newblock In {\em ECCV}, 2014.

\bibitem{liu2016deepfashion}
Ziwei Liu, Ping Luo, Shi Qiu, Xiaogang Wang, and Xiaoou Tang.
\newblock Deepfashion: Powering robust clothes recognition and retrieval with
  rich annotations.
\newblock In {\em CVPR}, 2016.

\bibitem{liu2015deep}
Ziwei Liu, Ping Luo, Xiaogang Wang, and Xiaoou Tang.
\newblock Deep learning face attributes in the wild.
\newblock In {\em ICCV}, 2015.

\bibitem{madras2018learning}
David Madras, Elliot Creager, Toniann Pitassi, and Richard Zemel.
\newblock Learning adversarially fair and transferable representations.
\newblock {\em arXiv preprint arXiv:1802.06309}, 2018.

\bibitem{miao2018comparison}
Zhongqi Miao, Kaitlyn~M Gaynor, Jiayun Wang, Ziwei Liu, Oliver Muellerklein,
  Mohammad~S Norouzzadeh, Alex McInturff, Rauri~CK Bowie, Ran Nathon, Stella~X.
  Yu, and Wayne~M. Getz.
\newblock A comparison of visual features used by humans and machines to
  classify wildlife.
\newblock {\em bioRxiv}, 2018.

\bibitem{mitchell2018model}
Margaret Mitchell, Simone Wu, Andrew Zaldivar, Parker Barnes, Lucy Vasserman,
  Ben Hutchinson, Elena Spitzer, Inioluwa~Deborah Raji, and Timnit Gebru.
\newblock Model cards for model reporting.
\newblock {\em arXiv preprint arXiv:1810.03993}, 2018.

\bibitem{munkhdalai2017meta}
Tsendsuren Munkhdalai and Hong Yu.
\newblock Meta networks.
\newblock {\em arXiv preprint arXiv:1703.00837}, 2017.

\bibitem{oh2016deep}
Hyun Oh~Song, Yu Xiang, Stefanie Jegelka, and Silvio Savarese.
\newblock Deep metric learning via lifted structured feature embedding.
\newblock In {\em CVPR}, 2016.

\bibitem{ouyang2016factors}
Wanli Ouyang, Xiaogang Wang, Cong Zhang, and Xiaokang Yang.
\newblock Factors in finetuning deep model for object detection with long-tail
  distribution.
\newblock In {\em CVPR}, 2016.

\bibitem{qi2018low}
Hang Qi, Matthew Brown, and David~G Lowe.
\newblock Low-shot learning with imprinted weights.
\newblock In {\em CVPR}, 2018.

\bibitem{qiao2018few}
Siyuan Qiao, Chenxi Liu, Wei Shen, and Alan Yuille.
\newblock Few-shot image recognition by predicting parameters from activations.
\newblock In {\em CVPR}, 2018.

\bibitem{ravi2016optimization}
Sachin Ravi and Hugo Larochelle.
\newblock Optimization as a model for few-shot learning.
\newblock In {\em ICLR}, 2017.

\bibitem{reed2001pareto}
William~J Reed.
\newblock The pareto, zipf and other power laws.
\newblock {\em Economics letters}, 2001.

\bibitem{ren2018incremental}
Mengye Ren, Renjie Liao, Ethan Fetaya, and Richard~S Zemel.
\newblock Incremental few-shot learning with attention attractor networks.
\newblock {\em arXiv preprint arXiv:1810.07218}, 2018.

\bibitem{sabour2017dynamic}
Sara Sabour, Nicholas Frosst, and Geoffrey~E Hinton.
\newblock Dynamic routing between capsules.
\newblock In {\em NIPS}, 2017.

\bibitem{salakhutdinov2011learning}
Ruslan Salakhutdinov, Antonio Torralba, and Josh Tenenbaum.
\newblock Learning to share visual appearance for multiclass object detection.
\newblock In {\em CVPR}, 2011.

\bibitem{santoro2016meta}
Adam Santoro, Sergey Bartunov, Matthew Botvinick, Daan Wierstra, and Timothy
  Lillicrap.
\newblock Meta-learning with memory-augmented neural networks.
\newblock In {\em ICML}, 2016.

\bibitem{savinov2018episodic}
Nikolay Savinov, Anton Raichuk, Rapha{\"e}l Marinier, Damien Vincent, Marc
  Pollefeys, Timothy Lillicrap, and Sylvain Gelly.
\newblock Episodic curiosity through reachability.
\newblock {\em arXiv preprint arXiv:1810.02274}, 2018.

\bibitem{scheirer2013toward}
Walter~J Scheirer, Anderson de Rezende~Rocha, Archana Sapkota, and Terrance~E
  Boult.
\newblock Toward open set recognition.
\newblock {\em TPAMI}, 2013.

\bibitem{schmidhuber1992learning}
J{\"u}rgen Schmidhuber.
\newblock Learning to control fast-weight memories: An alternative to dynamic
  recurrent networks.
\newblock {\em Neural Computation}, 1992.

\bibitem{schmidhuber1993neural}
J{\"u}rgen Schmidhuber.
\newblock A neural network that embeds its own meta-levels.
\newblock In {\em ICNN}, 1993.

\bibitem{shen2016relay}
Li Shen, Zhouchen Lin, and Qingming Huang.
\newblock Relay backpropagation for effective learning of deep convolutional
  neural networks.
\newblock In {\em ECCV}, 2016.

\bibitem{snell2017prototypical}
Jake Snell, Kevin Swersky, and Richard Zemel.
\newblock Prototypical networks for few-shot learning.
\newblock In {\em NIPS}, 2017.

\bibitem{van2017devil}
Grant Van~Horn and Pietro Perona.
\newblock The devil is in the tails: Fine-grained classification in the wild.
\newblock {\em arXiv preprint arXiv:1709.01450}, 2017.

\bibitem{vaswani2017attention}
Ashish Vaswani, Noam Shazeer, Niki Parmar, Jakob Uszkoreit, Llion Jones,
  Aidan~N Gomez, {\L}ukasz Kaiser, and Illia Polosukhin.
\newblock Attention is all you need.
\newblock In {\em NIPS}, 2017.

\bibitem{vinyals2016matching}
Oriol Vinyals, Charles Blundell, Tim Lillicrap, and Daan Wierstra.
\newblock Matching networks for one shot learning.
\newblock In {\em NIPS}, 2016.

\bibitem{wang2017non}
Xiaolong Wang, Ross Girshick, Abhinav Gupta, and Kaiming He.
\newblock Non-local neural networks.
\newblock {\em arXiv preprint arXiv:1711.07971}, 2017.

\bibitem{wang2018low}
Yu-Xiong Wang, Ross Girshick, Martial Hebert, and Bharath Hariharan.
\newblock Low-shot learning from imaginary data.
\newblock {\em arXiv preprint arXiv:1801.05401}, 2018.

\bibitem{wang2016learning}
Yu-Xiong Wang and Martial Hebert.
\newblock Learning to learn: Model regression networks for easy small sample
  learning.
\newblock In {\em ECCV}, 2016.

\bibitem{wang2017learning}
Yu-Xiong Wang, Deva Ramanan, and Martial Hebert.
\newblock Learning to model the tail.
\newblock In {\em NIPS}, 2017.

\bibitem{wen2016discriminative}
Yandong Wen, Kaipeng Zhang, Zhifeng Li, and Yu Qiao.
\newblock A discriminative feature learning approach for deep face recognition.
\newblock In {\em ECCV}, 2016.

\bibitem{yang2018learning}
Flood Sung~Yongxin Yang, Li Zhang, Tao Xiang, Philip~HS Torr, and Timothy~M
  Hospedales.
\newblock Learning to compare: Relation network for few-shot learning.
\newblock In {\em CVPR}, 2018.

\bibitem{zeiler2014visualizing}
Matthew~D Zeiler and Rob Fergus.
\newblock Visualizing and understanding convolutional networks.
\newblock In {\em ECCV}, 2014.

\bibitem{zemel2013learning}
Rich Zemel, Yu Wu, Kevin Swersky, Toni Pitassi, and Cynthia Dwork.
\newblock Learning fair representations.
\newblock In {\em ICML}, 2013.

\bibitem{zhang2017range}
Xiao Zhang, Zhiyuan Fang, Yandong Wen, Zhifeng Li, and Yu Qiao.
\newblock Range loss for deep face recognition with long-tailed training data.
\newblock In {\em CVPR}, 2017.

\bibitem{zhou2018places}
Bolei Zhou, Agata Lapedriza, Aditya Khosla, Aude Oliva, and Antonio Torralba.
\newblock Places: A 10 million image database for scene recognition.
\newblock {\em TPAMI}, 2018.

\bibitem{zhu2014capturing}
Xiangxin Zhu, Dragomir Anguelov, and Deva Ramanan.
\newblock Capturing long-tail distributions of object subcategories.
\newblock In {\em CVPR}, 2014.

\bibitem{zhu2016we}
Xiangxin Zhu, Carl Vondrick, Charless~C Fowlkes, and Deva Ramanan.
\newblock Do we need more training data?
\newblock {\em IJCV}, 2016.

\end{thebibliography}
}

\clearpage

\appendix
\addcontentsline{toc}{section}{Appendices}
\section*{Appendices}

In this supplementary material, we provide details omitted in the main text including:
\begin{itemize}[leftmargin=*]
\vspace{-6pt}
\item Section~\ref{sec:explanation}: intuitive explanation of our approach (Sec. 1 ``Introduction'' of the main paper.)
\vspace{-6pt}
\item Section~\ref{sec:fairness}: relation to fairness analysis (Sec. 2 ``Related Work'' of the main paper.)
\vspace{-6pt}
\item Section~\ref{sec:methodology}: more methodology details (Sec. 3 ``Approach'' of the main paper.)
\vspace{-6pt}
\item Section~\ref{sec:experimental}: detailed experimental setup (Sec. 4 ``Experiments'' of the main paper.)
\vspace{-6pt}
\item Section~\ref{sec:visualization}: additional visualization of our approach (Sec. 4.3 ``Further Analysis'' of the main paper.)
\end{itemize}

\section{Intuitive Explanation of Our Approach}
\label{sec:explanation}

\begin{figure}[h]
  \centering
  \includegraphics[width=0.45\textwidth]{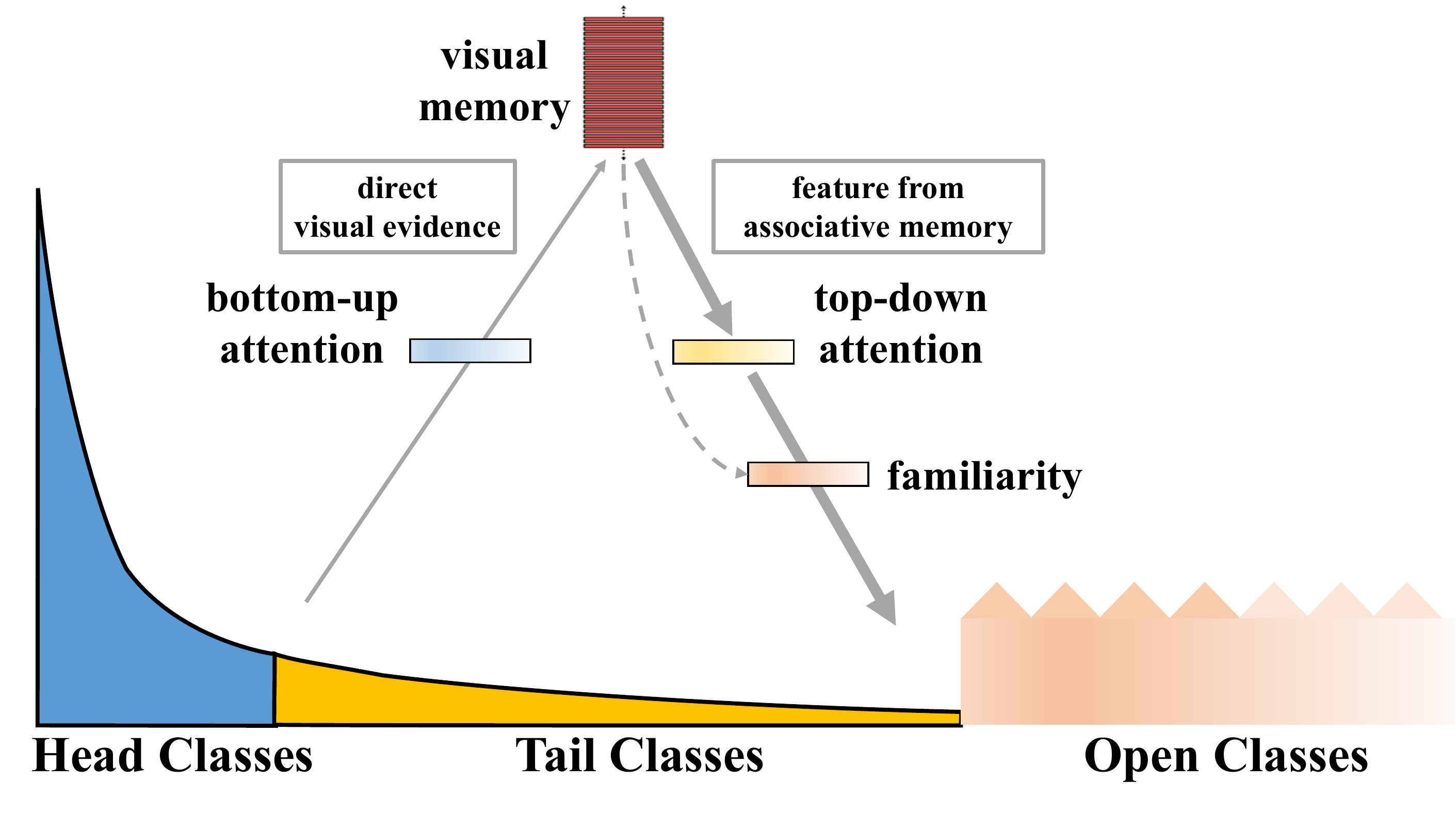}
  \vspace{-6pt}
  \caption{Intuition explanation of our approach.}
  \label{fig:intuition}
\end{figure}

In this section, we give an intuitive explanation of our approach that tackles the problem open long-tail recognition.
From the perspective of knowledge gained from observation (\ie training set), head classes, tail classes and open classes form a continuous spectrum as illustrated in Fig.~\ref{fig:intuition}.

\begin{table}[h]
    \centering
    \resizebox{0.5\textwidth}{!}{%
    \begin{tabular}{c|c|c}
    \Xhline{1pt}
    {\bf Direct + Memory Feature} & {\bf Modulated Attention} & {\bf Reachability Module} \\ \hline
    Transfer knowledge & Maintain discrimination & Deal with open classes \\
    between head/tail classes & between head/tail classes & \\
    \Xhline{1pt}
    \end{tabular}}
    \vspace{-6pt}
    \caption{The effects of each component in our approach.}
    \label{tab:effect}
\end{table}

Firstly, we obtain a \emph{visual memory} by aggregating the knowledge from both head and tail classes.
Then the visual concepts stored in the memory are infused back as associated memory feature to enhance the original direct feature.
It can be understood as using induced knowledge (\ie memory feature) to assist the direct observation (\ie direct feature).
We further learn a \emph{concept selector} to control the amount and type of memory feature to be infused.
Since head classes already have abundant direct observation, only a small amount of memory feature is infused for them.
On the contrary, tail classes suffer from scarce observation, the associated visual concepts in memory feature are extremely beneficial.
Finally, we calibrate the confidence of open classes by calculating their \emph{reachability} to the obtained visual memory.
In this way, we provide a comprehensive treatment to the full spectrum of head, tail and open classes, improving the performance on all categories.
To summarize, the effects of each component in our approach are listed in Table~\ref{tab:effect}.

\section{Relation to Fairness Analysis}
\label{sec:fairness}

The open long-tail recognition proposed in our work also has an intrinsic relationship to fairness analysis~\cite{dwork2012fairness, zemel2013learning, madras2018learning, mitchell2018model, anne2018women}. Their key differences are listed in Table~\ref{tab:differences}.
On the problem setting side, both open long-tail recognition and fairness analysis aim to tackle the imbalance existed in real-world data.
Open long-tail recognition focuses on the longtail-ness in both known and unknown categories while fairness analysis deals with the bias in sensitive attributes such as male/female and white/black.

On the methodology side, both open long-tail recognition and fairness analysis aim to learn transferable representations.
Open long-tail recognition optimizes for the overall accuracy of all categories while fairness analysis optimizes for several attribute-wise criteria.
The preliminary results in Table~\ref{tab:benchmark_megaface} demonstrates that our proposed dynamic meta-embedding is also a promising solution to fairness analysis.

\begin{table}[h]
    \footnotesize
    \centering
    \begin{tabular}{l|c|c}
    \Xhline{1pt}
    ~{\bf Problem}~ & ~{\bf Imbalanced Asp.}~ & ~{\bf Optimization Obj.}~ \\ \hline \hline
    fairness analysis & sensitive attributes  & attribute-wise criteria \\ \hline
    open long-tail recog. & categories  & acc. on all categories \\
    \Xhline{1pt}
    \end{tabular}
    \caption{Key differences between fairness analysis and open long-tail recognition. ``asp.'' stands for aspects while ``obj.'' stands for objectives.}
    \label{tab:differences}
\end{table}

\begin{figure*}[t]
  \centering
  \includegraphics[width=0.9\textwidth]{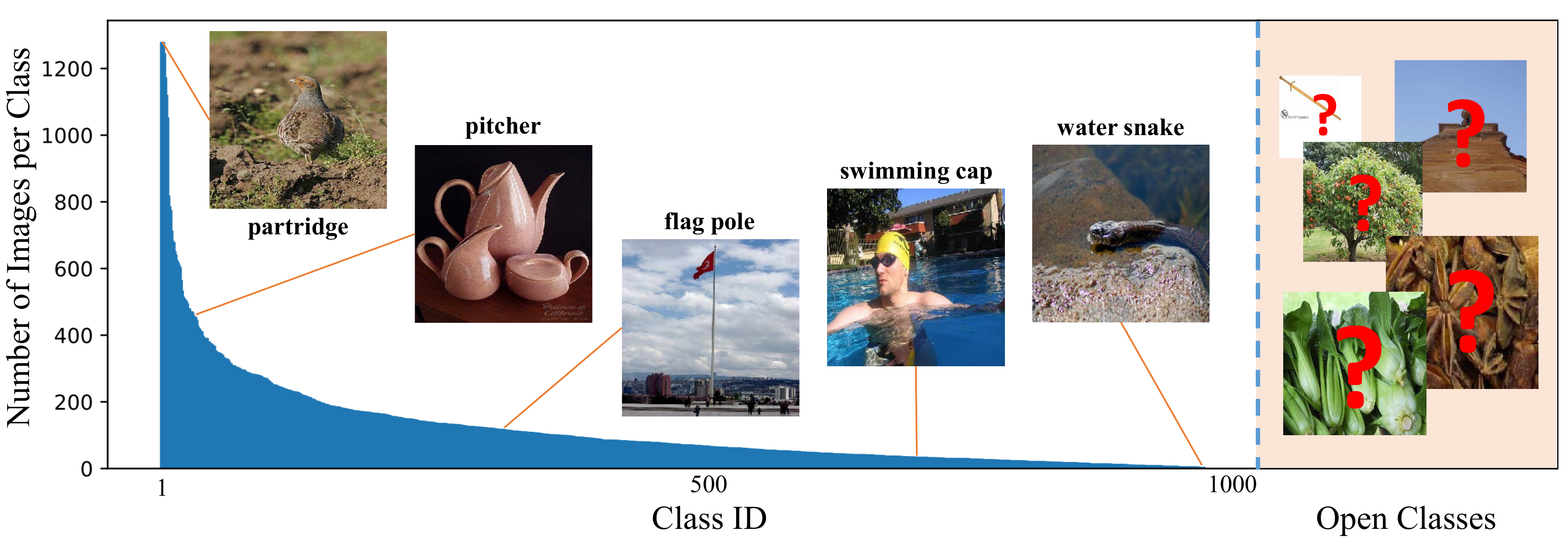}
  \vspace{-10pt}
  \caption{The dataset statistics of ImageNet-LT.}
  \label{fig:dataset_imagenet}
  \vspace{-10pt}
\end{figure*}

\section{More Methodology Details}
\label{sec:methodology}

\vspace{2pt}
\noindent
\textbf{Notation Summary.}
We summarize the notations used in the paper in Table~\ref{tab:notation}.

\begin{table}[h]
    \small
    \centering
    \begin{tabular}{l|c}
    \Xhline{1pt}
    ~{\bf Notation}~ & ~{\bf Meaning}~  \\ \hline \hline
    $x$ & input image \\ \hline
    $y$ & category label \\ \hline
    $f$ & the original feature map \\ \hline
    $f^{att}$ & feature map after modulated attention \\ \hline
    $F(\cdot)$ & feature extractor \\ \hline
    $\phi(\cdot)$ & classifier \\ \hline
    $c_{i}$ & discriminative centroid \\ \hline
    $G$ & local graph \\ \hline
    $M$ & visual memory \\ \hline
    $v^{direct}$ & direct feature \\ \hline
    $v^{memory}$ & memory feature \\ \hline
    $o$ & hallucinated coefficients from visual memory \\ \hline
    $e$ & concept selector \\ \hline
    $\gamma$ & confidence calibrator \\ \hline
    $v^{meta}$ & dynamic meta-embedding \\ 
    \Xhline{1pt}
    \end{tabular}
    \vspace{-6pt}
    \caption{Summary of notations.}
    \label{tab:notation}
\end{table}

\begin{figure}[h]
  \centering
  \includegraphics[width=0.45\textwidth]{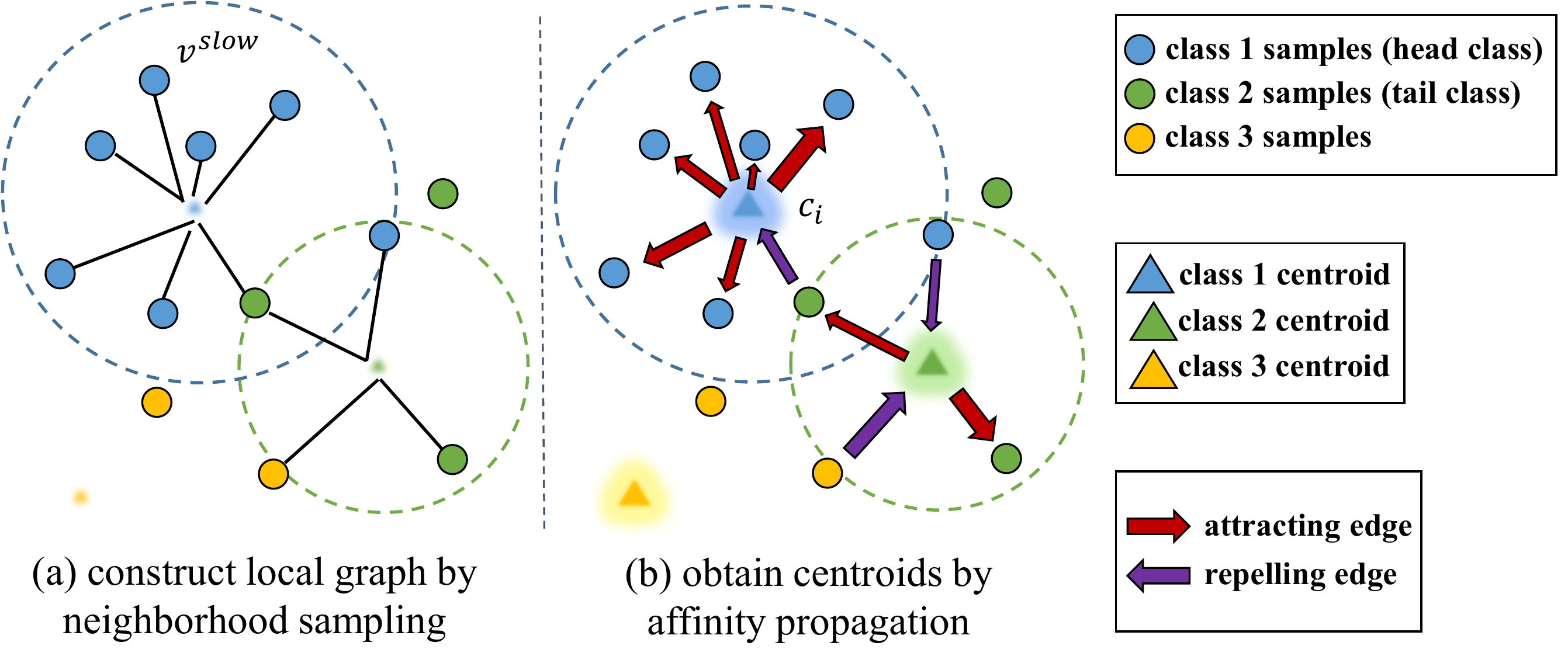}
  \caption{The discriminative centroids constitute our visual memory, which are obtained with two iterative steps, neighborhood sampling and affinity propagation.}
  \label{fig:centroid}
\end{figure}

\vspace{2pt}
\noindent
\textbf{Obtaining Discriminative Centroids.}
The step-by-step procedure for obtaining discriminative centroids $\{c_{i}\}_{i=1}^{K}$ is further illustrated in Fig.~\ref{fig:centroid}.

\vspace{2pt}
\noindent
\textbf{Detailed Loss Functions.}
Here we elaborate the two loss functions $L_{CE}$ and $L_{LM}$ described in Eqn. 7 in the main paper.
Specifically, $L_{CE}$ is the cross-entropy loss between dynamic meta-embedding $v^{meta}_{n}$ and the ground truth category label $y_{n}$:
\begin{equation}
\begin{split}
L_{CE}(v^{meta}_{n}, y_{n}) &= y_{n}\log(\phi(v^{meta}_{n})) \\
&+ (1- y_{n})\log(1 - \phi(v^{meta}_{n})),
\end{split}
\end{equation}
where $\phi(\cdot)$ is the cosine classifier described in Eqn. 6 in the main paper.
Next we introduce the large margin loss $L_{LM}$ between the embedding $v^{meta}_{n}$ and the centroids $\{c_{i}\}_{i=1}^{K}$:
\begin{equation}
\begin{split}
L_{LM}(v^{meta}_{n}, \{c_{i}\}_{i=1}^{K}) &= \max(0, \sum_{i = y_{n}} \|v^{meta}_{n} - c_{i} \| \\
&- \sum_{i \neq y_{n}} \|v^{meta}_{n} - c_{i} \| + m),
\end{split}
\end{equation}
where $m$ is the margin and we set it as $5.0$ in our experiments. With this formulation, we minimize the distance between each embedding and the centroid of its group and meanwhile maximize the distance between the embedding and the centroids it does not belong to.

\section{Experimental Setup}
\label{sec:experimental}

\subsection{Open Long-Tail Dataset Preparation}

\begin{figure*}[t]
  \centering
  \includegraphics[width=0.9\textwidth]{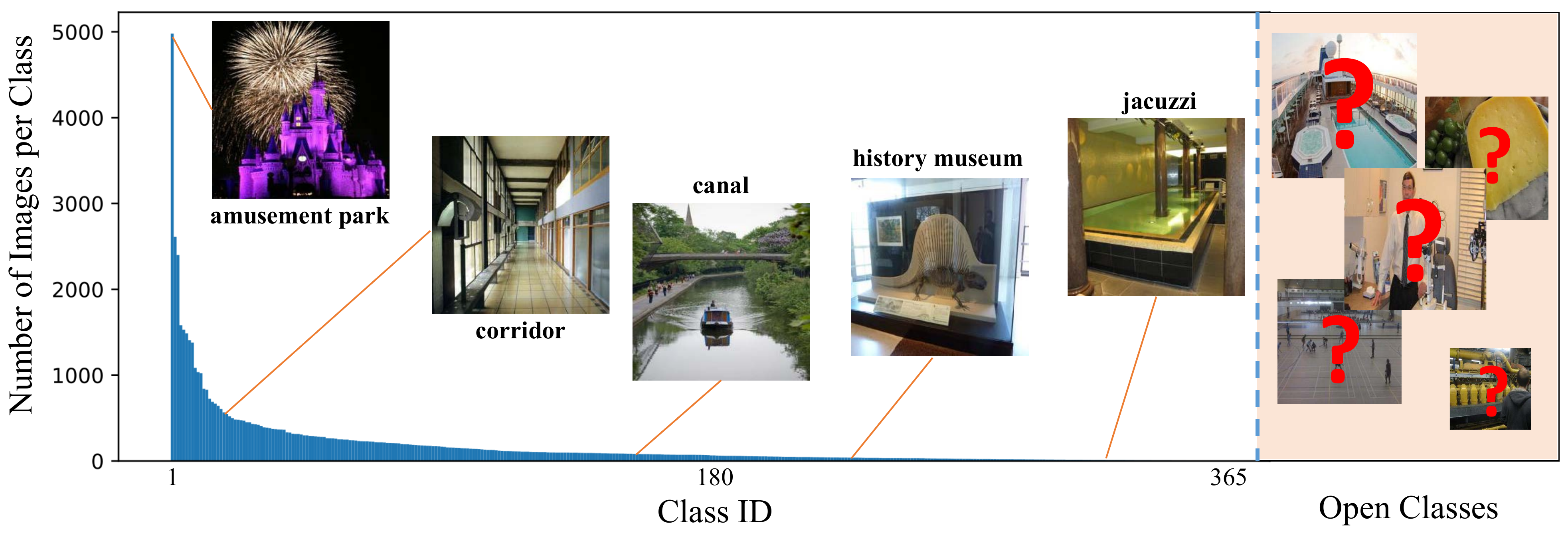}
  \vspace{-6pt}
  \caption{The dataset statistics of Places-LT.}
  \label{fig:dataset_places}
  \vspace{-10pt}
\end{figure*}

\begin{figure*}[t]
  \centering
  \includegraphics[width=0.9\textwidth]{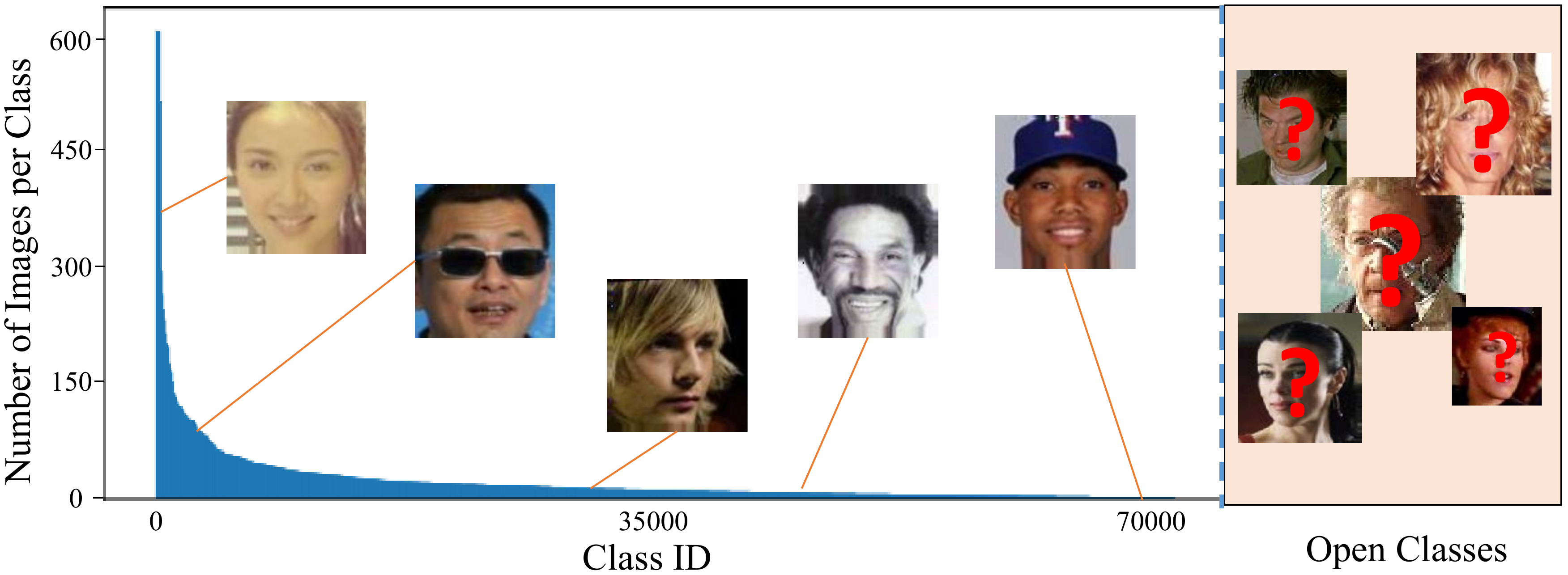}
  \vspace{-2pt}
  \caption{The dataset statistics of MS1M-LT.}
  \label{fig:dataset_ms1m}
  \vspace{-10pt}
\end{figure*}

\vspace{2pt}
\noindent
\textbf{ImageNet-LT.} 
The training data set was generated using a Pareto distribution~\cite{reed2001pareto} with a power value $\alpha$=6 and 1,280$\sim$5 images per class from the 1000 classes of ImageNet dataset. Images were randomly selected based on the distribution values of each class. The classes were sorted following the benchmark proposed by Bharath \& Girshick \cite{hariharan2017low}, where the 1000 classes were randomly split into 389 base classes and 611 novel classes. The first 389 largest classes in ImageNet-LT are the same as the base classes in the benchmark, and the rest 611 classes are the same as the novel classes. We randomly selected 20 training images per class from the origin training set as validation set. The original validation set of ImageNet was used as testing set in this paper. The dataset specifications are shown in Fig.~\ref{fig:dataset_imagenet}. 

\vspace{2pt}
\noindent
\textbf{Places-LT.} 
The training data set was generated similarly to ImageNet-LT using a Pareto distribution with a power value $\alpha$=6 and 4,980$\sim$5 images per class from the 365 classes of Places-365-standard data set. We used the distribution order of Places-365-challenge data set (which is imbalanced) to sort the training data classes. We also randomly selected 20 images per class from the original training set as validation set. The original validation set of Places-365 was used as testing set in this paper. The dataset specifications are shown in Fig.~\ref{fig:dataset_places}. 

\vspace{2pt}
\noindent
\textbf{MS1M-LT.}
This dataset was generated from a large-scale face recognition dataset, named MS1M-ArcFace. The original dataset contains about 5.8M images with 85K identities. To create a long-tail version, we sampled images for each identity with a probability proportional to the image numbers of each identity. It results in 887.5K images and 74.5K identities, with a long-tail distribution.

For the evaluation set, MegaFace is one of the largest face recognition benchmarks. It contains 3,530 images from FaceScrub dataset as a probe set and 1M images as a gallery set. The identification task is to find top-1 nearest image from the 1M gallery for each sample in the probe set. Then the identification rate is the mean of hit rates. Since the identities in training set and testing set are non-overlapped, we adopt an indirect way to partition the testing set into subsets with different shots. We approximate the pseudo occurrences of each test sample by counting the number of the similar (similarity greater than $0.7$) training samples. 
The similarity is calculated as the feature distance produced by a state-of-the-art face recognition system~\cite{deng2018arcface}.
Apart from many-shot, few-shot and one-shot subset, we also define a zero-shot subset, for which we cannot find similar samples in the training set.
The dataset specifications are shown in Fig.~\ref{fig:dataset_ms1m}.

\vspace{2pt}
\noindent
\textbf{SUN-LT.} 
We used the same training and testing data set as provided by \cite{wang2017learning}, where there were 1,132$\sim$1 images per class in the training set and 40 images per class in the testing set. We randomly selected 5 images from un-used training data as our validation set.

\subsection{Data Pre-processing}

All the images were firstly resized to $256 \times 256$. During training, the images were randomly cropped to $224 \times 224$, then augmented with random horizontal flip at probability $p=0.5$ and random color jitter on brightness, contrast, and saturation with jitter factor of 0.4. During validation and testing, images were center cropped to $224 \times 224$ without further augmentation.

\begin{figure*}
  \centering
  \includegraphics[width=1.0\textwidth]{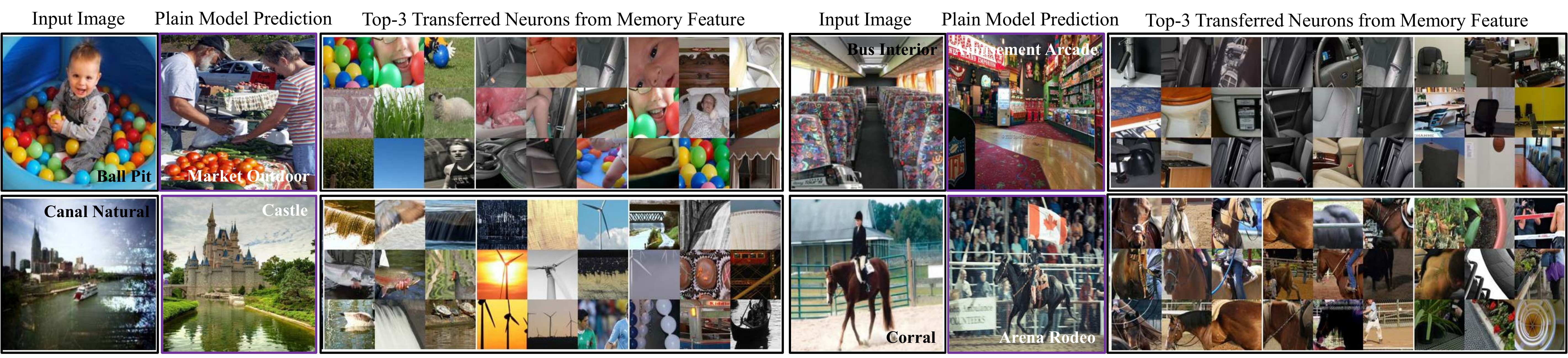}
  \vspace{-16pt}
  \caption{Examples of the infused visual concepts from memory feature in Places-LT.}
  \label{fig:concept_places}
\end{figure*}

\begin{figure*}
  \centering
  \includegraphics[width=1.0\textwidth]{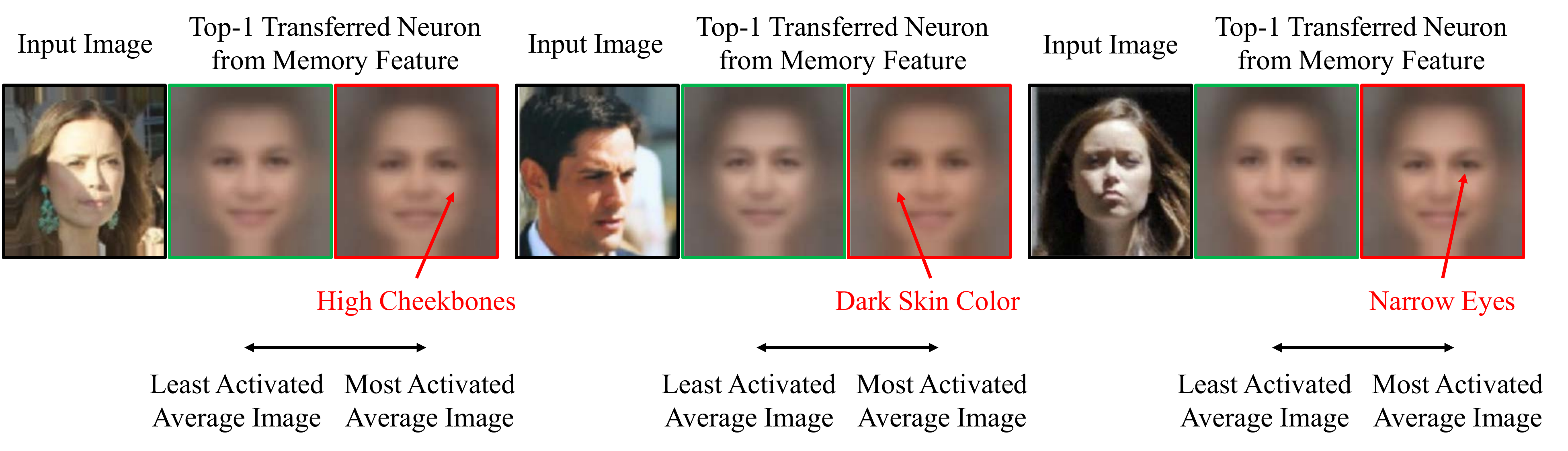}
  \vspace{-16pt}
  \caption{Examples of the infused visual concepts from memory feature in MS1M-LT.}
  \label{fig:concept_ms1m}
\end{figure*}

\subsection{Training Details}


\vspace{2pt}
\noindent
\textbf{ImageNet-LT.}
The feature extractor model used in the experiments on ImageNet-LT was a ResNet-10 model initialized from scratch (i.e., random initialization). All different classifiers were also initialized from scratch. Some major hyper-parameters can be found in Table~\ref{tab:parameters}.

\vspace{2pt}
\noindent
\textbf{Places-LT \& SUN-LT.}
We used a two-stage training protocol following~\cite{gidaris2018dynamic} when conducting experiments on both Places-LT and SUN-LT. (1) In the first stage, we used the ImageNet pre-trained ResNet-152 feature model with a dot-product classifier to fine-tune on the training data of Places-LT and SUN-LT. (2) In the second stage, we used the Places-LT/SUN-LT pre-trained model as our feature model and freezed the convolutional weights. Finally we fine-tuned the classifiers initialized from scratch to produce the experimental results. Some major hyper-parameters can be found in Table~\ref{tab:parameters}.

\vspace{2pt}
\noindent
\textbf{MS1M-LT.}
We used the ImageNet pre-trained ResNet-50 with a linear classifier and cross-entropy loss to train the  face recognition model.
Some major hyper-parameters can be found in Table~\ref{tab:parameters}.

\begin{table}[h]
    \footnotesize
    \centering
    \begin{tabular}{l|c|c|c}
    \Xhline{1pt}
    ~{\bf Dataset}~ & ~{\bf Initial LR.}~ & ~{\bf Epoch}~ & ~{\bf LR. Schedule}~ \\ \hline \hline
    ImageNet-LT & 0.1  & 30 & drop 10\% every 10 epochs \\ \hline
    Places-LT & 0.01 & 30 & drop 10\% every 10 epochs \\ \hline
    MS1M-LT & 0.01 & 30 & drop 10\% every 10 epochs \\
    \Xhline{1pt}
    \end{tabular}
    \vspace{-6pt}
    \caption{The major hyper-parameters used in our experiments. ``LR.'' stands for learning rate.}
    \label{tab:parameters}
\end{table}

\subsection{Evaluation Protocols}

\vspace{2pt}
\noindent
\textbf{Top-1 Classification Accuracy.}
For ImageNet-LT, Places-LT, and SUN-LT, since the testing sets are balanced, the top-1 classification accuracy are calculated as the mean accuracy over all close-set categories with the contamination of open classes.
All open classes are regared as one unknown class.
Predictions of data are obtained as the classes with the highest $softmax$ probabilities. 


\vspace{2pt}
\noindent
\textbf{F-measure.}
Following \cite{bendale2016towards}, the F-measure ($F$) is calculated as $2$ times the product of precision ($p$) and recall ($r$) divided by the sum of $p$ and $r$:
\begin{equation}
    F = 2 \cdot\frac{p \cdot r}{p + r}.
\end{equation}
$p$ is calculated as true positive ($T_p$, defined as correct predictions on the closed testing set) over the sum of $T_p$ and false positive ($F_p$, defined as incorrect predictions on closed testing set):
\begin{equation}
    p = \frac{T_p}{T_p + F_p}.
\end{equation}
$r$ is calculated as $T_p$ over the sum of $T_p$ and false negative ($F_n$, defined as number of images from the open set that are predicted as known categories):
\begin{equation}
    r = \frac{T_p}{T_p + F_n}.
\end{equation}






\section{More Visualization}
\label{sec:visualization}

\vspace{2pt}
\noindent
\textbf{Memory Feature in Places-LT.}
We visualize the memory feature in Places-LT similarly to ImageNet-LT as described in Sec. 4.3 in the main paper.
Examples of the infused visual concepts from memory feature in Places-LT are presented in Fig.~\ref{fig:concept_places}.
We observe that memory feature encodes discriminative visual traits for the underlying scene.

\vspace{2pt}
\noindent
\textbf{Memory Feature in MS-1M.}
Following~\cite{liu2015deep}, we visualize the memory feature in MS1M-LT by contrasting the least activated average image and the most activated average image of the top firing neuron.
From Fig.~\ref{fig:concept_ms1m}, we observe that memory feature in MS1M-LT infuses several identity-related attributes (\eg ``high cheekbones'', ``dark skin color'' and ``narrow eyes'') for precise recognition.

\end{document}